\documentclass[10pt,twocolumn,letterpaper]{article}

\usepackage{wacv}
\makeatletter
\@namedef{ver@everyshi.sty}{}
\makeatother
\usepackage{tikz}
\usepackage{times}
\usepackage{epsfig}
\usepackage{graphicx}
\usepackage{amsmath}
\usepackage{amssymb}
\usepackage{cite}
\usepackage{svg}
\usepackage{subcaption}
\usepackage[export]{adjustbox}

\DeclareMathOperator*{\argmax}{arg\,max}
\newcommand\numberthis{\addtocounter{equation}{1}\tag{\theequation}}

\def\wacvPaperID{1018} 

\wacvfinalcopy 
\ifwacvfinal
\def\assignedStartPage{9876} 
\fi

\ifwacvfinal
\usepackage[breaklinks=true,bookmarks=false]{hyperref}
\else
\usepackage[pagebackref=true,breaklinks=true,colorlinks,bookmarks=false]{hyperref}
\fi

\ifwacvfinal
\else
\fi

\begin{document}

\title{Foreground color prediction through inverse compositing}

\author{Sebastian Lutz\\
V-SENSE\\
Trinity College Dublin\\
{\tt\small lutzs@scss.tcd.ie}
\and
Aljosa Smolic\\
V-SENSE\\
Trinity College Dublin\\
{\tt\small smolica@scss.tcd.ie}
}

\maketitle



\begin{abstract}
In natural image matting, the goal is to estimate the opacity of the foreground object in the image. This opacity controls the way the foreground and background is blended in transparent regions. In recent years, advances in deep learning have led to many natural image matting algorithms that have achieved outstanding performance in a fully automatic manner. However, most of these algorithms only predict the alpha matte from the image, which is not sufficient to create high-quality compositions. Further, it is not possible to manually interact with these algorithms in any way except by directly changing their input or output. We propose a novel recurrent neural network that can be used as a post-processing method to recover the foreground and background colors of an image, given an initial alpha estimation. Our method outperforms the state-of-the-art in color estimation for natural image matting and show that the recurrent nature of our method allows users to easily change candidate solutions that lead to superior color estimations.
\end{abstract}

\section{Introduction}
\begin{figure*}
    \centering
    \begin{subfigure}[t]{0.33\textwidth}
    \centering
        \includegraphics[width=\linewidth]{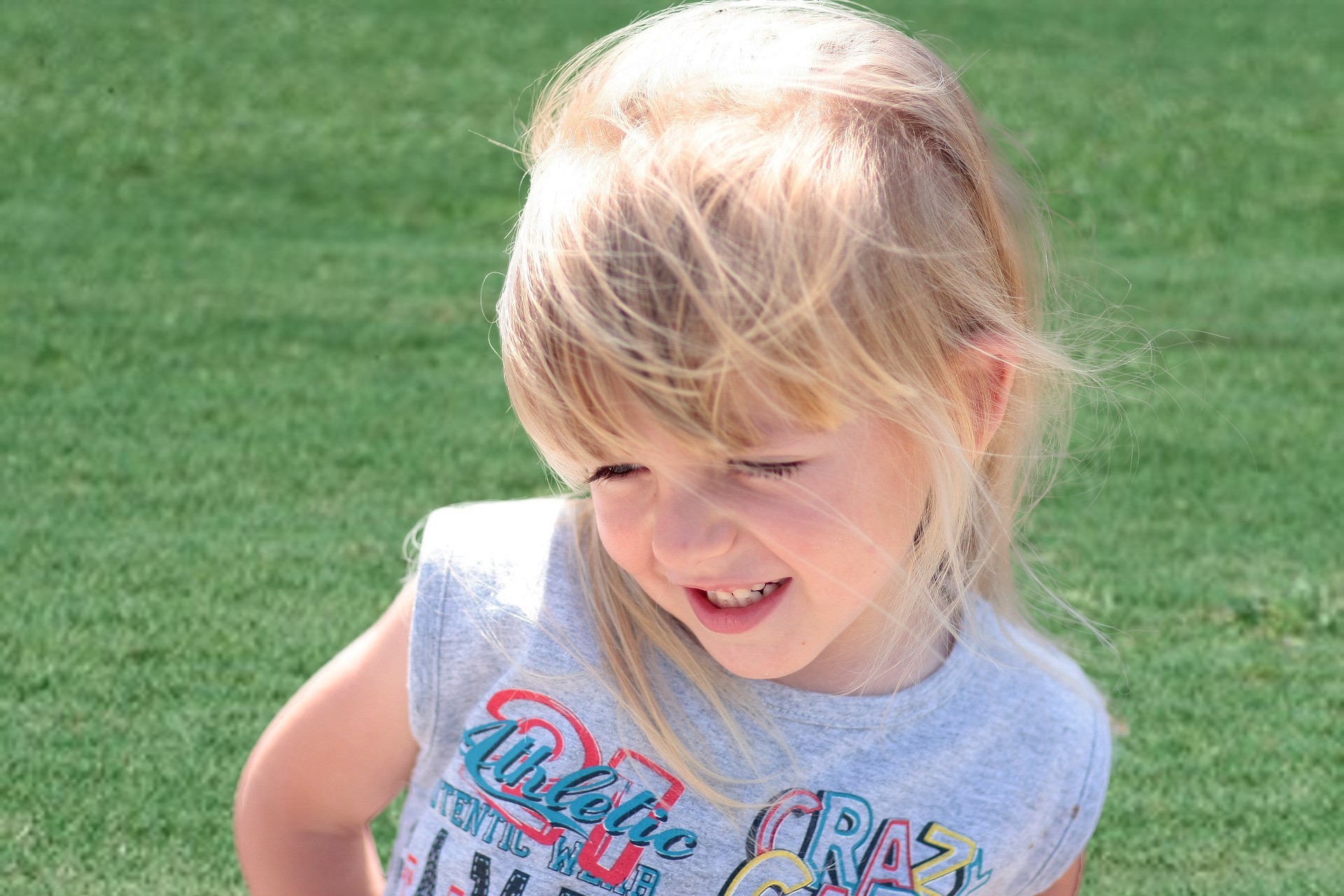}
    \end{subfigure}
    \hfill
    \begin{subfigure}[t]{0.33\textwidth}
    \centering
        \includegraphics[width=\linewidth]{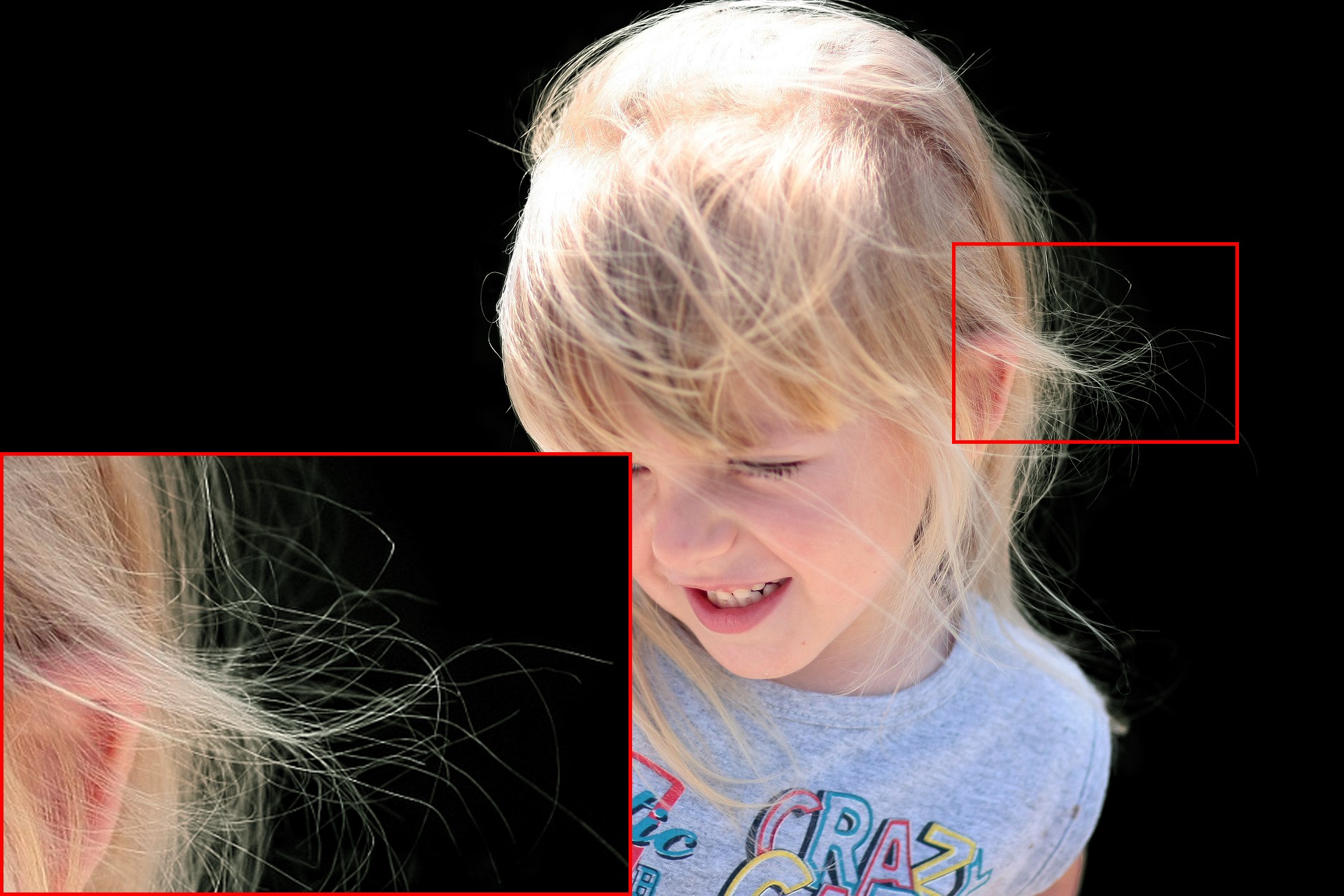}
    \end{subfigure}
    \hfill
    \begin{subfigure}[t]{0.33\textwidth}
    \centering
        \includegraphics[width=\linewidth]{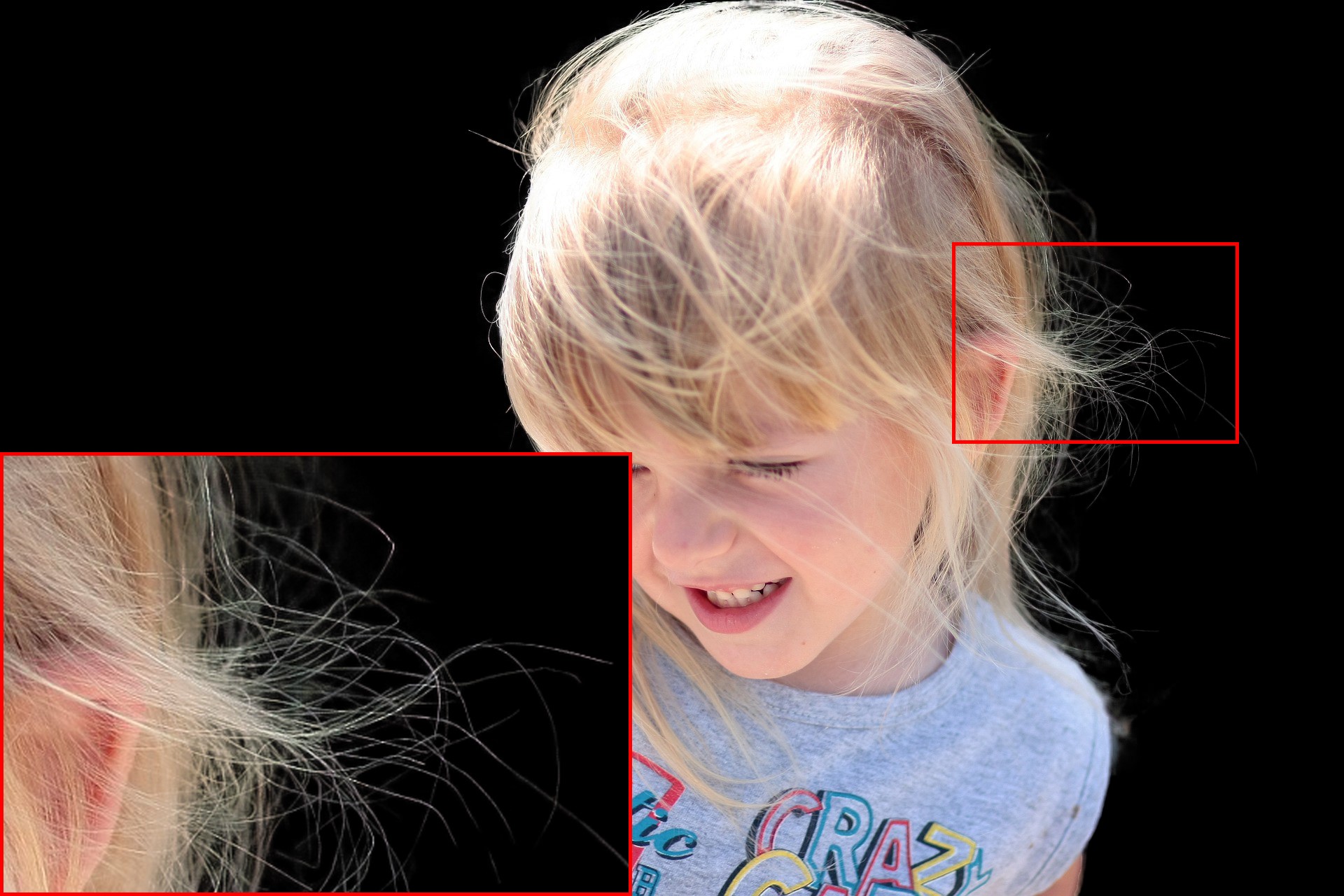}
    \end{subfigure}
    
    \caption{
    To the left: The input image. In the middle: The foreground extracted from the input image using the ground-truth alpha and composited onto a black background. To the right: The foreground extracted from our predicted foreground using the predicted alpha and composited onto a black background. As can be seen, even when using the ground-truth alpha, the green of the old background shines through in the new composition. This is not the case in our composition.}
    \label{fig:comp}
\end{figure*}

Natural image matting is one of the classical problems in computer vision and has been researched extensively in the past. The goal of this task is to predict the transition from a foreground object in the image to the background. As opposed to semantic segmentation, these transitions are not hard borders between the foreground and background, but rather values between $0$ and $1$ that denote the opacity of the foreground object. Mathematically, the input image $I$ is the combination of the foreground image $F$ and background image $B$ according to the compositing equation:
\begin{equation}
\label{eq:matting-equation}
    I_i = \alpha_i F_i + (1 - \alpha_i)B_i, \hspace{20pt} \alpha_i \in [0, 1],
\end{equation}
for every pixel $i$ in the image. The \textit{alpha matte} $\alpha_i$ that blends the foreground and background images indicates the opacity of the foreground object. This problem contains 7 unknown and only 3 known variables and is therefore severely ill-posed. To help with that, many matting algorithms use a trimap as additional input. The trimap is a rough segmentation into pure foreground, pure background and an unknown area that could contain any alpha value. This additional input is especially necessary for older affinity-based or sampling-based methods that estimate alpha values from image patches with known alpha values, i.e. patches in the predefined known areas designated by the trimap. However, even for state-of-the-art learning-based methods, the trimap proves useful to guide the algorithm and helps it focus on the desired area of the image.\\
Natural image matting algorithms are in use in various applications, such as image and video editing, as well as compositing and film post-production. However, current state-of-the-art algorithms have shortcomings when used in real-world scenarios. First of all, if the predicted alpha is used for extracting the foreground object from an image to composite a new one, the alpha alone is often not enough to create a high quality composite even with a perfect alpha prediction. This is due to the mixture of foreground and background colors in the transparent regions of the object. The background color of the original image bleeding through will be extracted as well and can lead to a disparity of colors in the new composite in many cases, as seen in figure \ref{fig:comp}. For high quality composites, it is therefore necessary to also estimate the foreground color of the object. The second problem with current matting methods is, that they do not allow for much user interaction if the prediction is not quite satisfactory. Often, the only tuning option for a user is to refine the input trimap, which may not be good enough. In high-quality studio environments for example, it is necessary that artists can easily refine predictions to a very high level without having to resort to manually touching up the prediction on a pixel-wise level.\\
In our approach, we aim to solve both of these issues by estimating not only the alpha value, but also the foreground and background colors of the input image and allowing more user interaction into the prediction process. We model our network as a recurrent inference machine (RIM) \cite{DBLP:journals/corr/PutzkyW17} that sits atop existing matting methods and is trained fully end-to-end with reconstruction and Wasserstein GAN losses \cite{DBLP:journals/corr/ArjovskyCB17, DBLP:conf/nips/GulrajaniAADC17}. Our contributions are therefore:
\begin{itemize}
    \item A fully automatic algorithm that can be added to any existing alpha prediction algorithm and that estimates the foreground and background colors for a given alpha, essentially solving the inverse compositing problem.
    \item Due to the recurrent nature of our algorithm, we allow users to update intermediate predictions to further guide the algorithm, which can lead to much better predictions through only a minor amount of manual work.
    \item We show through experiments that our color predictions substantially surpass the state-of-the-art. Especially our foreground color prediction leads to faithful new composites.
    \item Our method is small and lightweight and can process even high resolution images very fast.
\end{itemize}

    



\section{Related Work}
Traditionally, natural image matting methods could be classified into affinity-based and sampling-based methods. However, as with other computer vision tasks, deep learning methods have proven to be the state-of-the-art in natural image matting in recent years. Cho et al. \cite{DBLP:conf/eccv/ChoTK16} proposed a method that combines the predictions of two previous local and nonlocal matting algorithms and the RGB image as inputs to a convolutional neural network (CNN) that directly learns a mapping from these inputs to reconstructed alpha mattes. Due to the sparsity of training data, their network was quite small by current standards and only predicted $27 \times 27$ patches of the input. Recognizing this deficiency, Xu et al. \cite{DBLP:conf/cvpr/XuPCH17} made two contributions in their publication. They proposed an encoder-decoder network based on VGG16 \cite{DBLP:journals/corr/SimonyanZ14a} that could predict the alpha matting from a full resolution input image and trimap directly. To train this network, they also released a dataset containing $431$ high resolution images and their corresponding alpha, as well as a benchmark test set that contains a further 50 unique images. Due to the new availability of additional training data to fully train even larger CNNs, many other methods followed based on Xu et al. Lutz et al. \cite{DBLP:conf/bmvc/LutzAS18} proposed a generative adversarial network that leveraged the adversarial loss to improve on high-frequency structures in the alpha. Tang et al. \cite{DBLP:conf/cvpr/TangAOGA19} recognized that any alpha prediction would be more accurate if it had the actual foreground and background colors as input. They proposed a framework of three consecutive networks that would first predict the background colors in the unknown region, then predict the foreground colors based on the input image and the predicted background colors, and finally predict the alpha using the previous predictions. Hou et al. \cite{DBLP:conf/iccv/Hou019} created a network that simultaneously predicts the alpha and foreground color of the object. Cai et al. \cite{DBLP:journals/corr/abs-1909-04686} proposed a multitask autoencoder that disentangles the matting problem into two sub-tasks: The trimap adaption tasks that predicts the definitive trimap from a coarse input trimap and the alpha estimation part. They show improved performance due to more structural awareness. Li et al. \cite{DBLP:journals/corr/abs-2001-04069} proposes the use of a Guided Contextual Attention module in their network that directly propagates high-level opacity information based on learned low-level affinity. Finally Zhang et al. \cite{DBLP:conf/cvpr/ZhangGFRHBX19} developed a method that does not rely on an input trimap for their approach. They directly predict foreground and background classification maps from the input image and use this to predict the alpha of the foreground object.\\
Unrelated to the previously mentioned matting papers, Putzky et al. \cite{DBLP:journals/corr/PutzkyW17} proposed a novel network architecture to solve inverse problems on which we base our method. They introduce a recurrent inference machine that takes an initial guess for the solution as input and predicts an update step in each iteration. This allows them to solve several different image restoration tasks such as denoising and super-resolution.


\section{Method}
\begin{figure*}[t]
  \centering
  \centerline{\includegraphics[width=\textwidth]{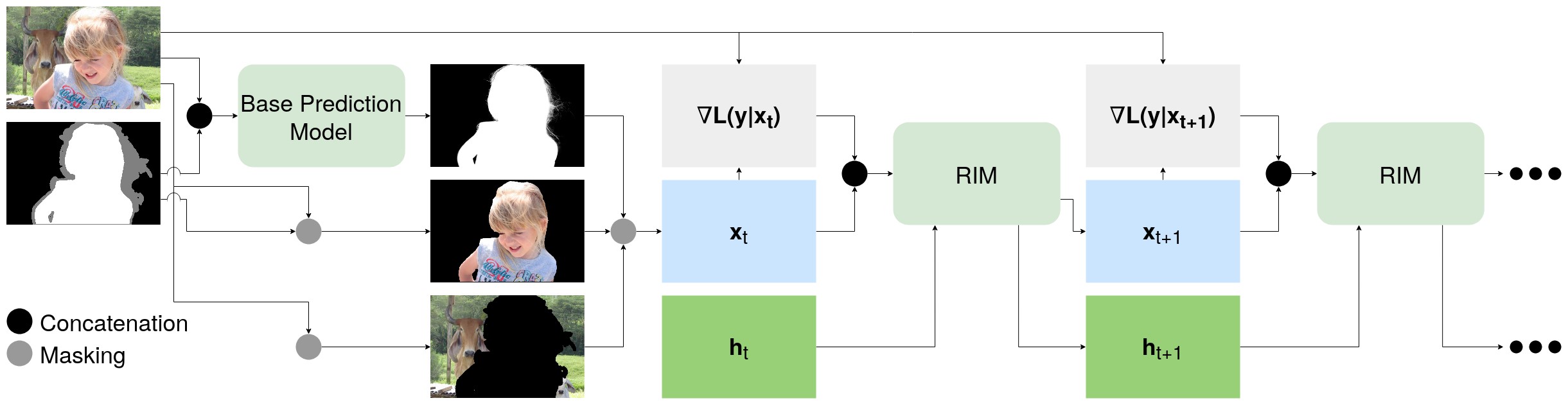}}
\caption{The Overall system. First, we do an initial alpha prediction using any prediction model. Next, we create initial foreground and background predictions, using either the trimap or the initial alpha prediction as a mask for the input image. All three initial predictions form the 7-channel $\mathbf{x}_t$. Given the original image $\mathbf{y}$ and $\mathbf{x}_t$, we calculate the gradient of the log-likelihood $\nabla L(\mathbf{y}|\mathbf{x}_t)$, which we concatenate to $\mathbf{x}_t$ to form the input of the RIM. The RIM then calculates and applies an update step $\Delta \mathbf{x}_{t+1}$ for $\mathbf{x}_t$ to form $\mathbf{x}_{t+1}$.}
\label{fig:system}
\end{figure*}

Our method aims to solve the inverse compositing problem by simultaneously estimating the alpha matte, as well as the foreground and background colors of a given image. Instead of trying to solve either of these problems from scratch, we rely on an initial guess for the solution of the alpha and leverage the correlation of the three problems to solve for the foreground and background colors in the process. As a result of this, our method can be seen as a post-processing method that can be used on any of the many methods that aim to predict the alpha matte of an image. Our method estimates the foreground and background colors, slightly refines the alpha, and additionally gives users the ability to easily refine the results further with easy manual user interaction as described in section \ref{sec:editing}.\\
To achieve this, we design a recurrent inference machine  \cite{DBLP:journals/corr/PutzkyW17} to solve the inverse of the forward model given in equation \ref{eq:matting-equation}. Traditionally, one way to achieve this would be to optimize the maximum a posteriori (MAP) solution, given a likelihood and prior as has been done by Bayesian Matting \cite{DBLP:conf/cvpr/ChuangCSS01} in the past:
\begin{align*}
    \argmax_{F,B,\alpha} P(F,B,\alpha|I) &= L(I|F,B,\alpha)\\
    &+ L(F) + L(B) + L(\alpha), \numberthis \label{eq:map}
\end{align*}
where $L(\cdot)$ is the \textit{log likelihood} $L(\cdot) = log P(\cdot)$, $\alpha$ is the alpha matte and F, B, I are the foreground, background and observed image colors respectively.\\
In contrast, a RIM as proposed by Putzky et al. \cite{DBLP:journals/corr/PutzkyW17} is a recurrent neural network (RNN) that is able to learn the iterative inference of the problem and implicitly learns the prior and inference procedure in the model parameters. Each state of the RIM includes the current solution, a hidden memory state and the gradient of the likelihood to the problem, which gives information about the generative process and indicates how good the current solution is. Given an observation $\mathbf{y}$ and a previous solution $\mathbf{x}_{t-1}$, the RIM calculates the gradient of the log-likelihood $\nabla L(\mathbf{y}|\mathbf{x}_{t-1})$ as an additional input to the network and predicts an update step $\Delta \mathbf{x}_t$ such that
\begin{equation}
    \mathbf{x}_{t} = \mathbf{x}_{t-1} + \Delta \mathbf{x}_{t},
\end{equation}
as can be seen in figure \ref{fig:system}. In this paper, the image $I$ is the observation $\mathbf{y}$ and the foreground, background and alpha $F, B, \alpha$ form the current solution $\mathbf{x}_t$.
The log-likelihood in our case is modeled by the difference between the color in the observed image and the color that would result from the composition of the predicted foreground, background, and alpha \cite{DBLP:conf/cvpr/ChuangCSS01}:
\begin{equation}
    \label{eq:log-likelihood}
    L(\mathbf{y}|\mathbf{x}) = L(I|F,B,\alpha) = - \frac{||I - \alpha F - (1 - \alpha)B||^2}{\sigma^2}.
\end{equation}
This corresponds to Gaussian probability distribution centered around $\bar{C} = \alpha F + (1 - \alpha) B$ with a standard deviation of $\sigma$. From this, the gradient of this log-likelihood is given by:
\begin{equation}
    \label{eq:gradient}
    \begin{aligned}
    \nabla L(\mathbf{y}|\mathbf{x}) &= \nabla L(I|F,B,\alpha) =
\begin{bmatrix} 
\dfrac{\partial L}{\partial F},
\dfrac{\partial L}{\partial B},
\dfrac{\partial L}{\partial \alpha}
\end{bmatrix}^T\\
    &= \begin{bmatrix}
    2 \alpha (I-\alpha F + B - \alpha B)\\
    (-2 + 2 \alpha) (I-\alpha F + B - \alpha B)\\
    ||(2F+2B)(I-\alpha F + B - \alpha B)||_1\\
    \end{bmatrix} * \frac{1}{\sigma^2}.
    \end{aligned}
\end{equation}
Since $\dfrac{\partial L}{\partial \alpha}$ is a sum across all three RGB channels, we abbreviate the term with $||\cdot||_1$.
As stated previously, our method serves as a post-processing for other alpha prediction methods. As such, we use the output of whichever alpha prediction method we use as a base as the initial guess for the alpha. For the initial foreground and background predictions, we use the original input image in all areas where the trimap gives known foreground/background regions respectively and zeros otherwise. If the base method does not need to use a trimap, we simply use the regions where the predicted alpha is purely foreground or purely background as mask.\\
The RIM then predicts updates for the current solution at every iteration, that consecutively lead to an ideal solution for all predictions. A loss is calculated for every iteration with the final loss being defined as:
\begin{equation}
    \label{eq:total-loss}
    \mathcal{L}^{total} = \sum_{t=1}^T w_t \mathcal{L}(\mathbf{x}_t, \mathbf{x}_{target}),
\end{equation}
where $T$ is the total number of iterations, $w_t$ is a positive weighting factor and $\mathbf{x}_{target}$ is the ground-truth. In this paper we set $T=5$ and $w_t = 1$ for all iterations, which we experimentally found to achieve the best results. Further details of our loss function are given in section \ref{sec:loss}. A visualization of the iterative process can be seen in the supplementary materials.

\subsection{Network architecture}
\begin{figure*}[t]
  \centering
  \centerline{\includegraphics[width=0.9\textwidth]{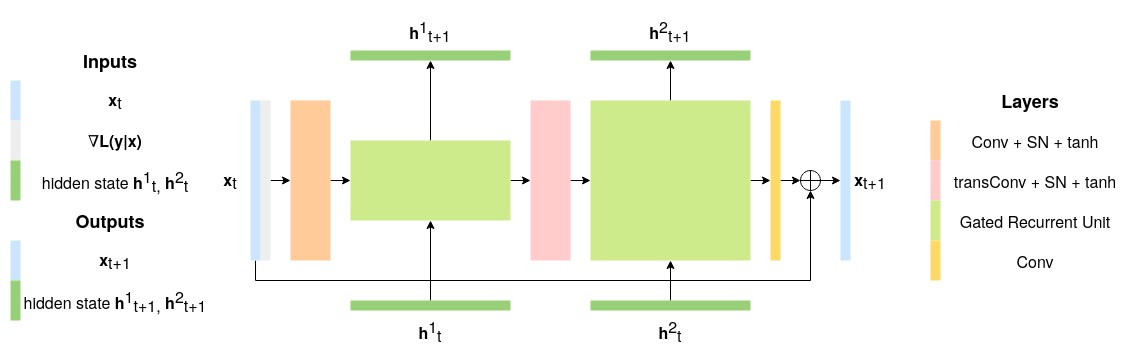}}

%
\caption{Network architecture. For each iteration, a candidate solution $\mathbf{x}_t$ serves as input, is concatenated with the gradient $\nabla L(\mathbf{y}|\mathbf{x})$ and put through the network to predict an update step, which is added to $\mathbf{x}_t$. Convolutions are normalized through spectral normalization (SN) and use tanh as activation function.}
\label{fig:network}
\end{figure*}

When designing our network architecture, we deliberately kept it small and lightweight. The full input of the network consists of the current solution $\mathbf{x}_t = [F,B,\alpha]^T$ and the gradient of the log-likelihood $\nabla L(I|F,B,\alpha)$, resulting in a full resolution feature map of $14$ channels in total. The first convolution layer of the network downsamples the input by a factor of $2$ and is followed by a gated recurrent unit (GRU). Following that, a transposed convolution layer upsamples the resulting feature maps back to the original resolution, again followed by a GRU. A final convolution layer serves as output layer and reduces the number of channels back to 7. With the exception of the last one, all convolution layers use spectral normalization \cite{DBLP:conf/iclr/MiyatoKKY18} and a tanh nonlinearity. The number of feature maps for the inner two convolutions is $32$ respectively and the number of feature maps for the hidden states of the GRUs is $128$. All convolution kernels are $3 \times 3$. The full network structure can be seen in figure \ref{fig:network}.\\
Our network only contains $1155680$ parameters, which easily fits even on mobile devices. However, the network operates on full resolution feature maps and propagates full resolution hidden states to the next iteration, which may use a lot of memory. This is not an issue, however, since the receptive field of the network is only $11 \times 11$. Due to this, even very high resolution images can be processed in smaller tiles with an overlap between tiles of only $11$ pixels.

\subsection{Loss Function}
\label{sec:loss}
Inspired by the success of generative inpainting networks \cite{DBLP:conf/cvpr/Yu0YSLH18, DBLP:journals/corr/abs-1806-03589}, we use the WGAN-GP loss \cite{DBLP:journals/corr/ArjovskyCB17, DBLP:conf/nips/GulrajaniAADC17}, combined with a $l_1$-based reconstruction loss to train our network.\\ Following previous methods \cite{DBLP:conf/bmvc/LutzAS18, DBLP:conf/cvpr/TangAOGA19, DBLP:journals/corr/abs-2001-04069, DBLP:conf/cvpr/XuPCH17, DBLP:journals/corr/abs-1909-04686}, we define the reconstruction loss only over the unknown region of the image. This leads to the reconstruction loss at iteration $t$:
\begin{equation}
    \mathcal{L}_1 = \frac{1}{|\mathcal{U}|} \sum_{i \in \mathcal{U}} |\mathbf{x}_t - \mathbf{x}_{target}|,
\end{equation}
where $\mathcal{U}$ is the unknown region of the trimap. Note that we do not calculate the loss independently for the alpha, foreground and background predictions. $\mathbf{x}_{target}$ is defined as the ground-truth foreground, background and alpha concatenated to a $7$-channel feature map and $\mathbf{x}_t$ as the update step predicted by the RIM at iteration $t$ added to the previous prediction: $\mathbf{x}_t = \mathbf{x}_{t-1} + \Delta \mathbf{x}_t$. \\
We further add the WGAN-GP loss to train the RIM with adversarial gradients and improve prediction accuracy. 

\subsection{Training details}
We train our network on the publicly available matting dataset published by Xu et al. \cite{DBLP:conf/cvpr/XuPCH17}. This dataset contains $431$ unique foreground images and corresponding alpha mattes. They further release another $50$ unique images that can be composited with predefined backgrounds to create the Composition-1k testing dataset.
To generate the initial alpha predictions during training, we use GCA-Matting \cite{DBLP:journals/corr/abs-2001-04069} as a base since they achieve the best performance on the Composition-1k testing set. To produce the initial foreground and background predictions, we use the given trimap to mask the input image in their respective known regions. Afterwards, all images are normalized to the range of $[-1, 1]$. Since the training dataset is small, we follow the improved data augmentation strategy of Li et al. \cite{DBLP:journals/corr/abs-2001-04069}. This strategy initially selects two random foreground images with a probability of $0.5$ and combines them to create a new foreground object with corresponding alpha. They follow this by resizing the image to $640 \times 640$ with a probability of $0.25$ to generate some images that contain nearly the whole image instead of random patches. Following this, a random affine transformation is applied to the image, which consists of random rotation, scaling, shearing and flipping. Afterwards, a trimap is generated by random dilation and a $512 \times 512$ patch is cropped from the image, centered around a random pixel in the unknown region of the trimap. Then, the image is converted to HSV space and random jitter is applied to hue, saturation and value. Finally, a random background image is selected from MS COCO \cite{DBLP:conf/eccv/LinMBHPRDZ14} and the input image is composited. Naturally, all applicable transformations are applied to the foreground image, alpha and trimap to keep them matching.\\
We use the Adam optimizer \cite{DBLP:journals/corr/KingmaB14} with a fixed learning rate of $10^{-4}$ and train for $100000$ iterations.

\section{Manual editing}
\label{sec:editing}
\begin{figure*}
    \centering
    \begin{adjustbox}{minipage=\textwidth,scale=1.0}
    \begin{subfigure}[t]{0.24\textwidth}
        \centering
        \includegraphics[width=\linewidth]{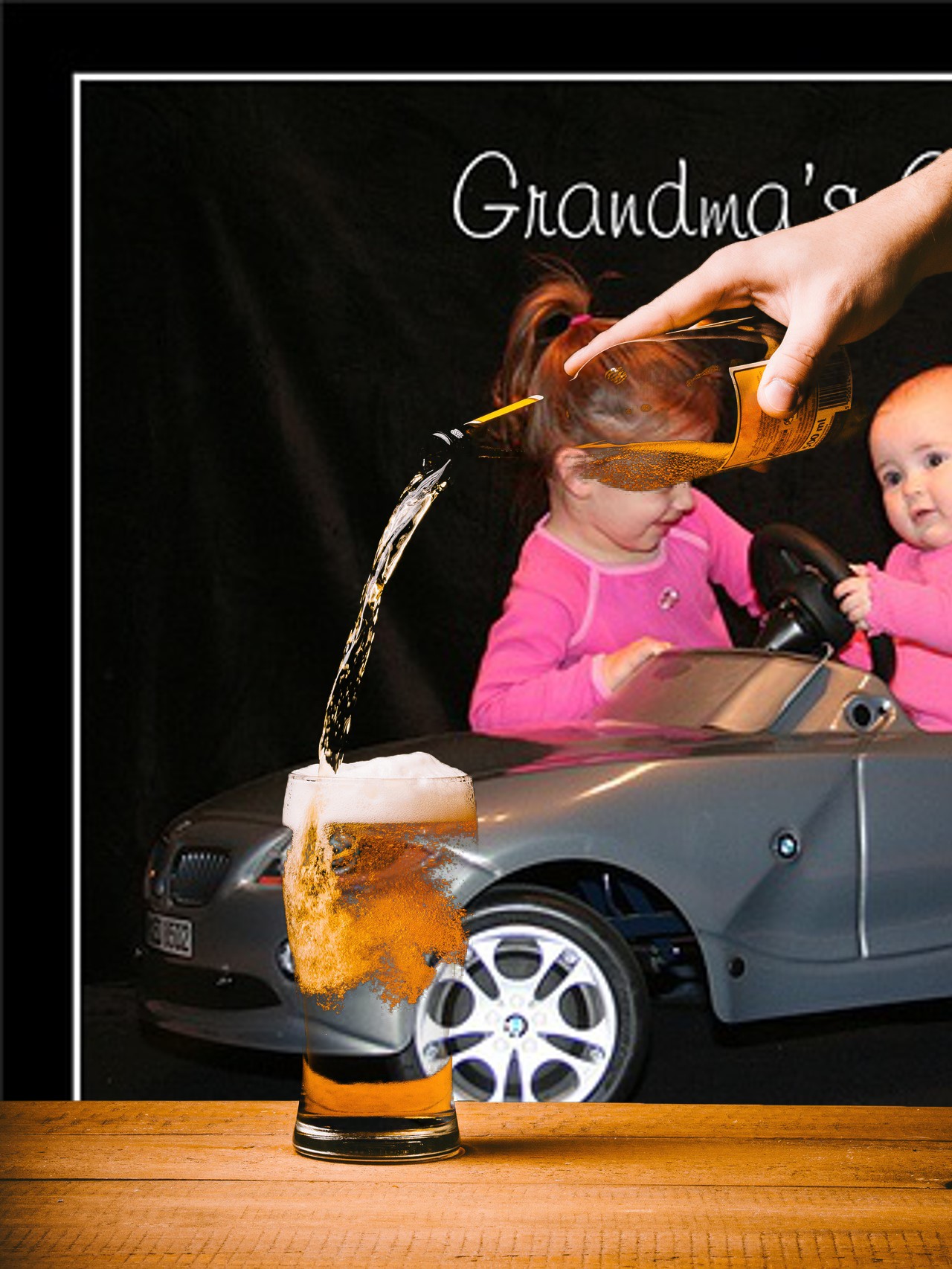} 
    \end{subfigure}
    \hfill
    \begin{subfigure}[t]{0.24\textwidth}
        \centering
        \includegraphics[width=\linewidth]{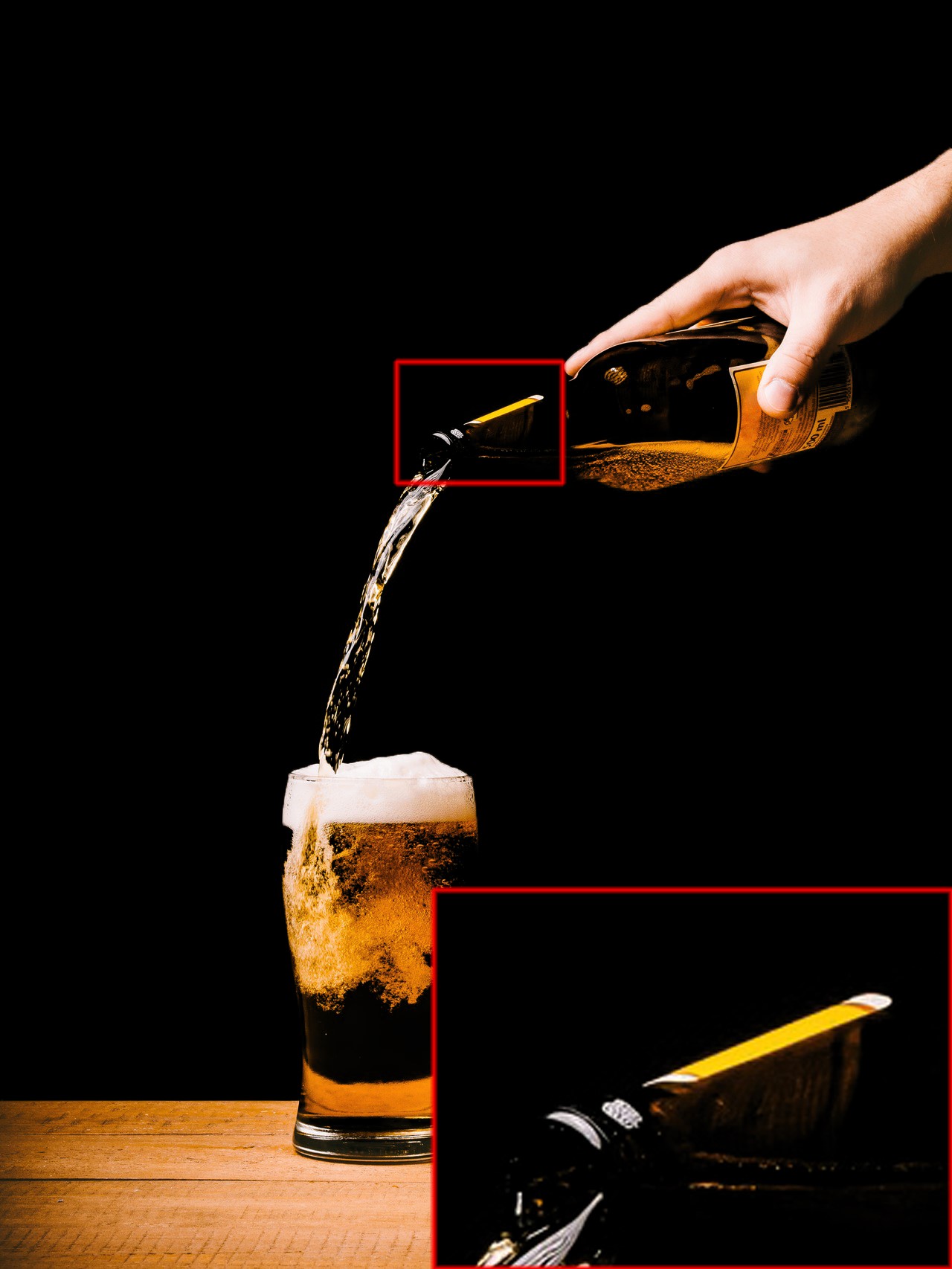} 
    \end{subfigure}
    \hfill
    \begin{subfigure}[t]{0.24\textwidth}
        \centering
        \includegraphics[width=\linewidth]{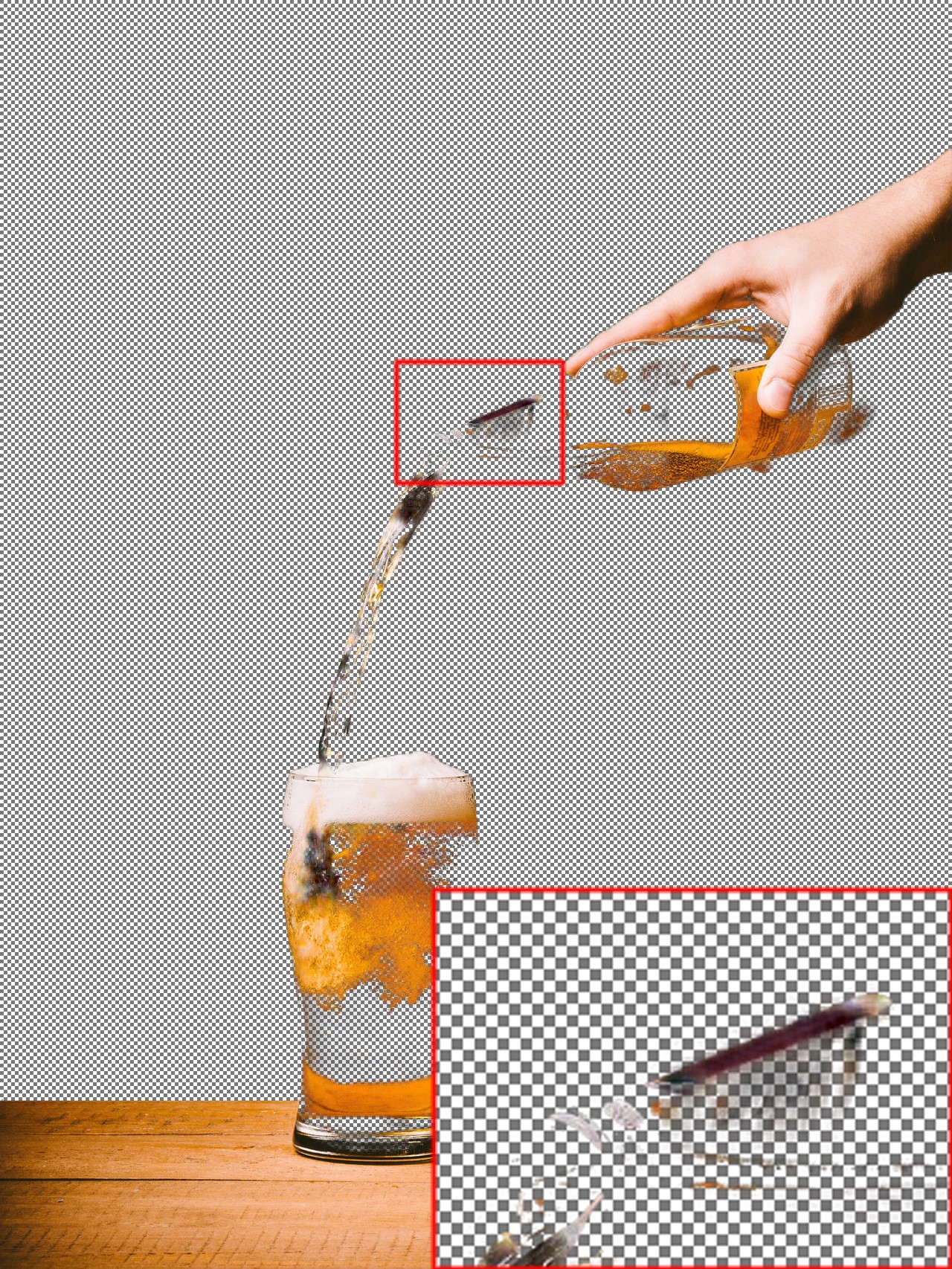} 
    \end{subfigure}
    \hfill
    \begin{subfigure}[t]{0.24\textwidth}
        \centering
        \includegraphics[width=\linewidth]{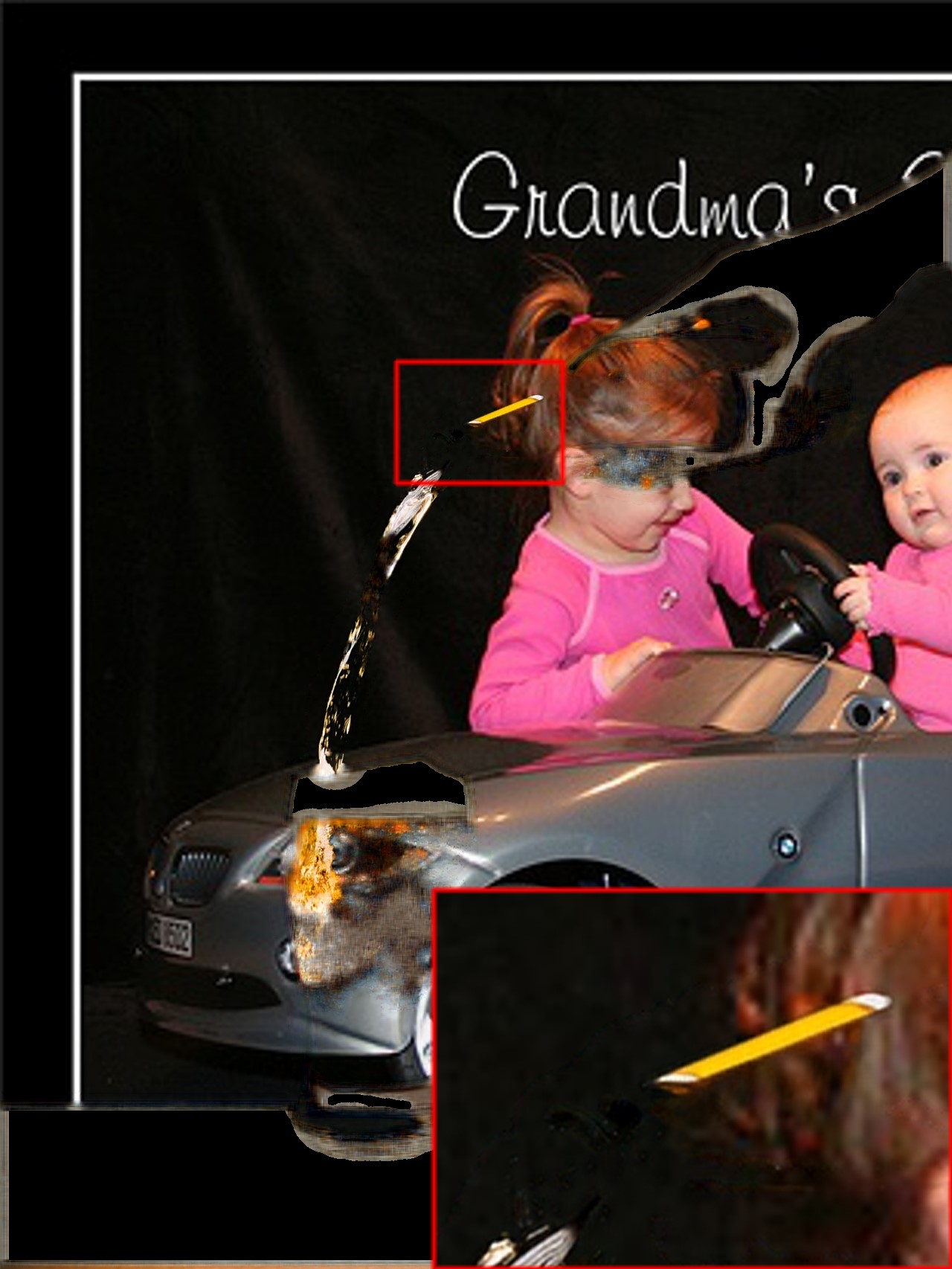} 
    \end{subfigure}
    \hfill
    \vspace{0.2cm}
    
    \begin{subfigure}[t]{0.24\textwidth}
        \centering
        \includegraphics[width=\linewidth]{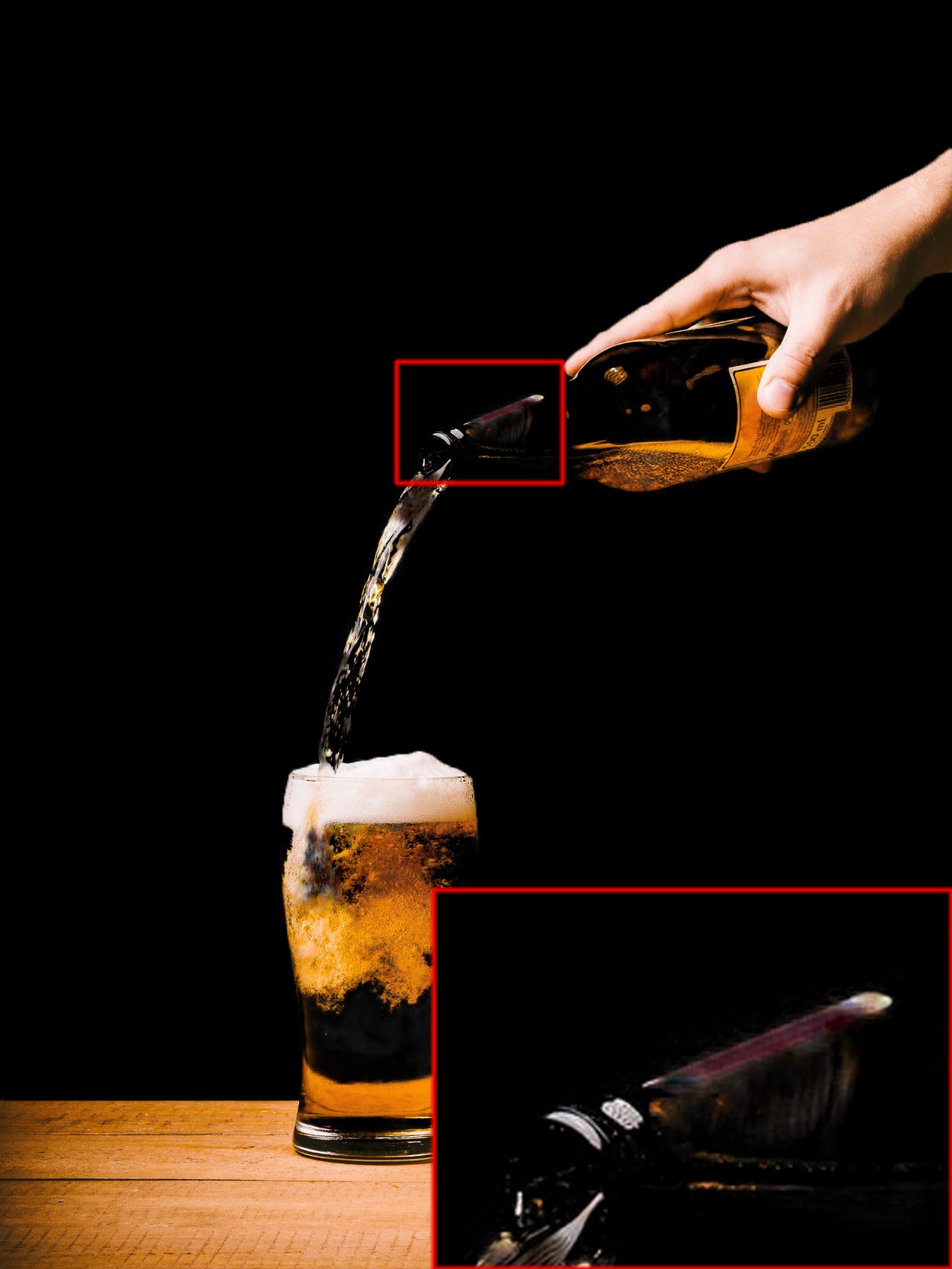} 
    \end{subfigure}
    \hfill
    \begin{subfigure}[t]{0.24\textwidth}
        \centering
        \includegraphics[width=\linewidth]{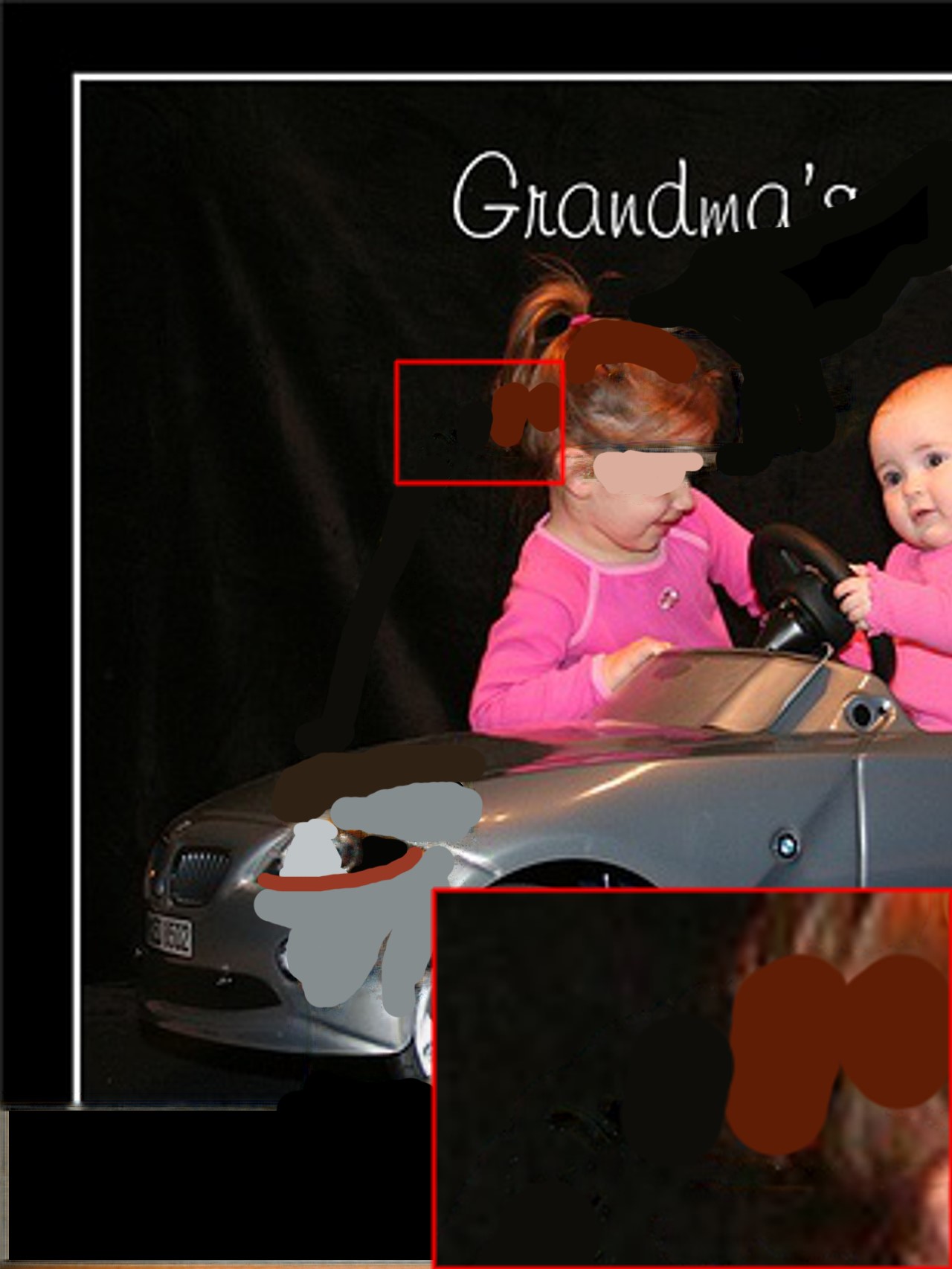} 
    \end{subfigure}
    \hfill
    \begin{subfigure}[t]{0.24\textwidth}
        \centering
        \includegraphics[width=\linewidth]{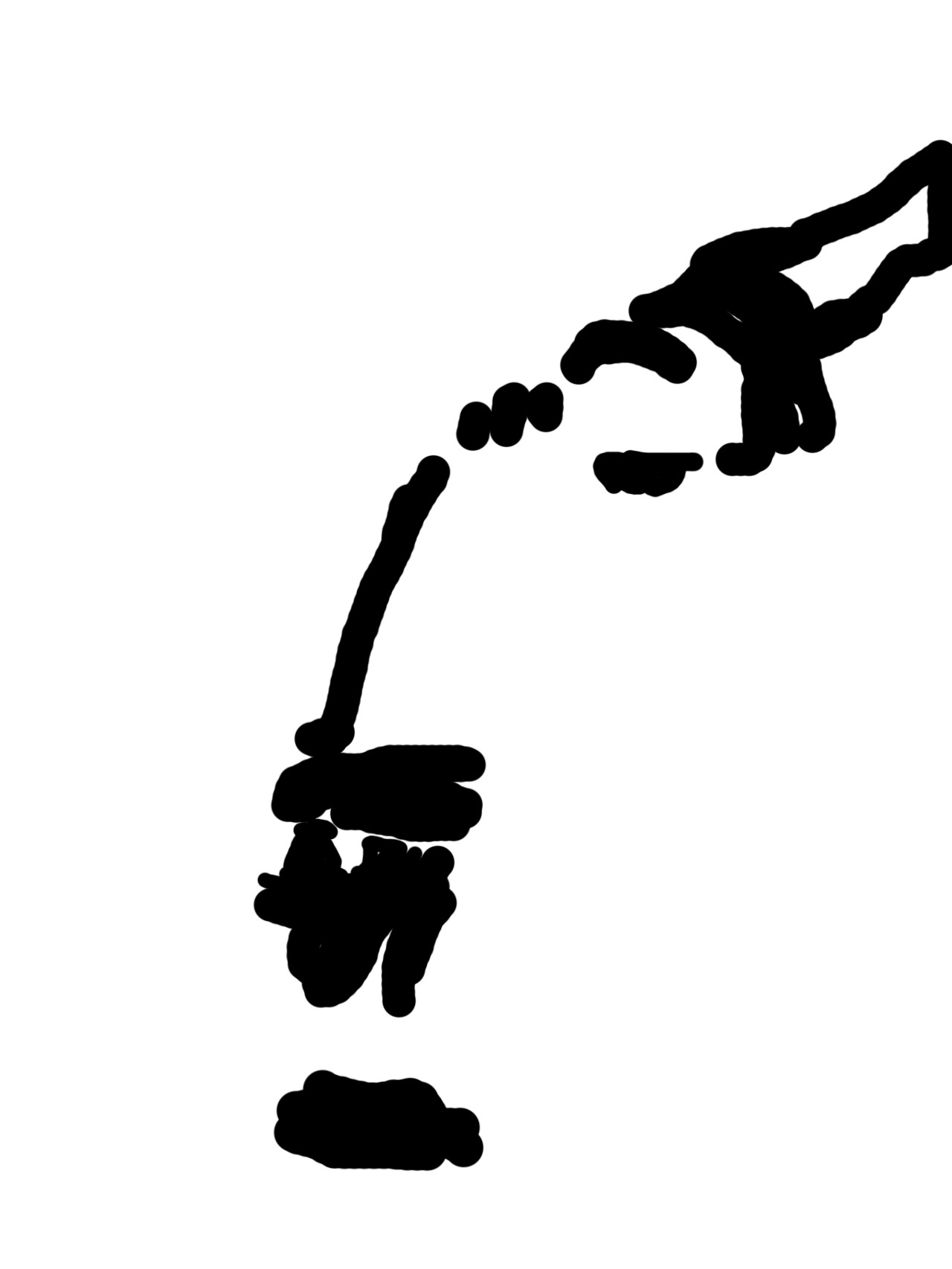} 
    \end{subfigure}
    \hfill
    \begin{subfigure}[t]{0.24\textwidth}
        \centering
        \includegraphics[width=\linewidth]{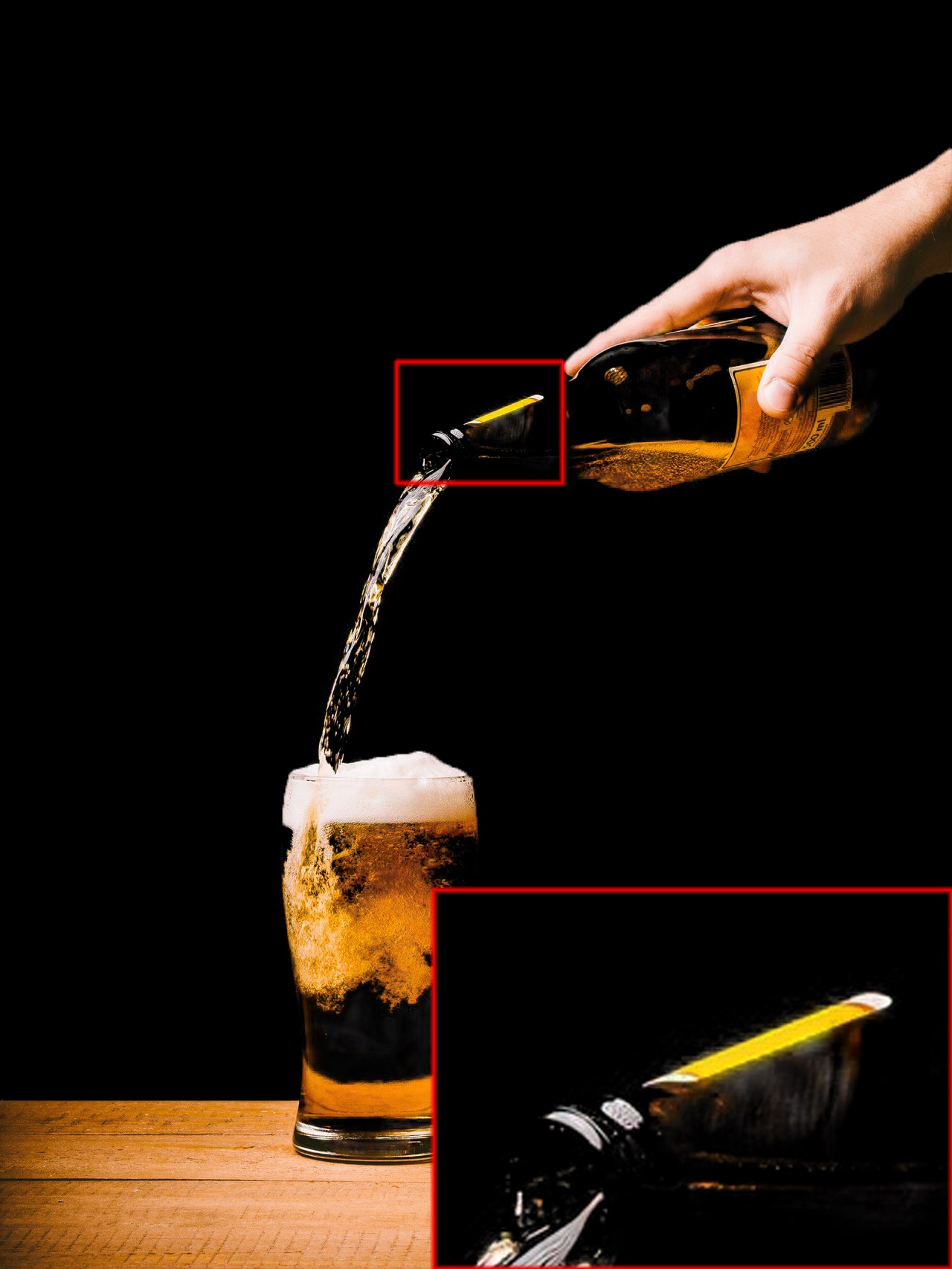} 
    \end{subfigure}
    \vspace{0.2cm}
    \end{adjustbox}
    
    \caption{Visualization of the manual editing process. Top row from left to right: Input image, ground-truth composition, predicted foreground, predicted background. Bottom row from left to right: Resulting composition, very rough edit to the background, corresponding editing mask and the composition resulting from manual edit. As can be seen, the color predictions in this example are bad and incorrectly color parts of the background with foreground colors and vice versa. After the rough manual edit to the background, the algorithm recovers the correct foreground color.}
    \label{fig:edit}
    
\end{figure*}

One of the main objectives of our method is to give users the optional ability to more directly influence the prediction process. Due to the recurrent nature of our method, this is easily achievable. During any step of the prediction, users can directly manually update any one of the foreground, background or alpha predictions through image editing software. Since all three predictions are fundamentally linked, changes in one of them can propagate to the other two. For example, a user may manually inpaint part of the background that the network struggles with and which may be easier to adjust for the user than directly changing either the foreground colors or the alpha. This change propagates to the foreground prediction and can lead to a much better foreground in the end. A detailed example of this can be seen in figure \ref{fig:edit}. To make sure that the network recognizes any direct modification of the predictions by the user, we additionally set the hidden states at the corresponding locations to $0$. The whole process is intuitive and can easily be implemented as part of any image processing software.

\section{Results}

\begin{figure*}
    \centering
    \begin{subfigure}[t]{0.19\textwidth}
        \centering
        \includegraphics[width=\linewidth]{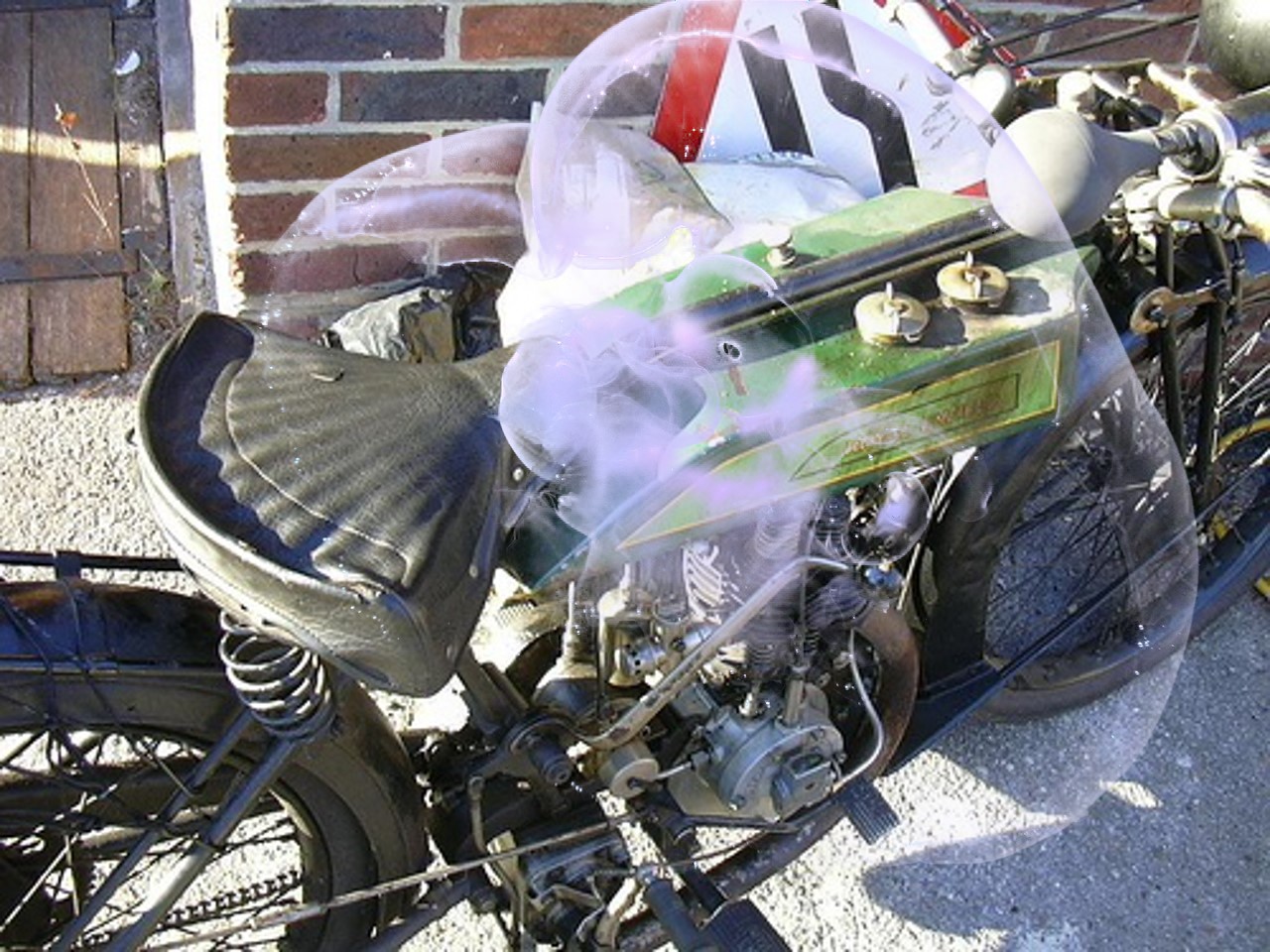} 
    \end{subfigure}
    \hfill
    \begin{subfigure}[t]{0.19\textwidth}
        \centering
        \includegraphics[width=\linewidth]{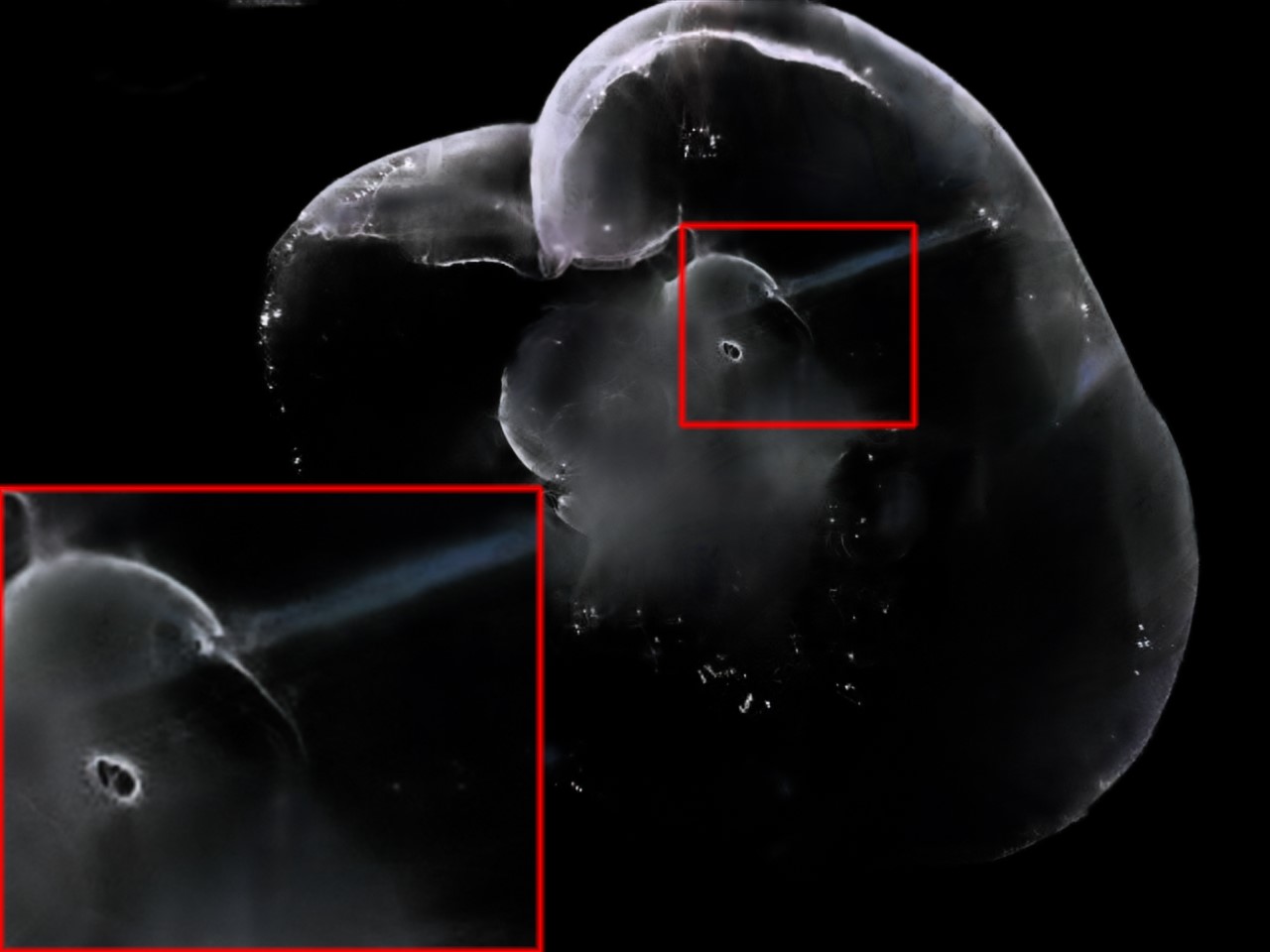} 
    \end{subfigure}
    \hfill
    \begin{subfigure}[t]{0.19\textwidth}
        \centering
        \includegraphics[width=\linewidth]{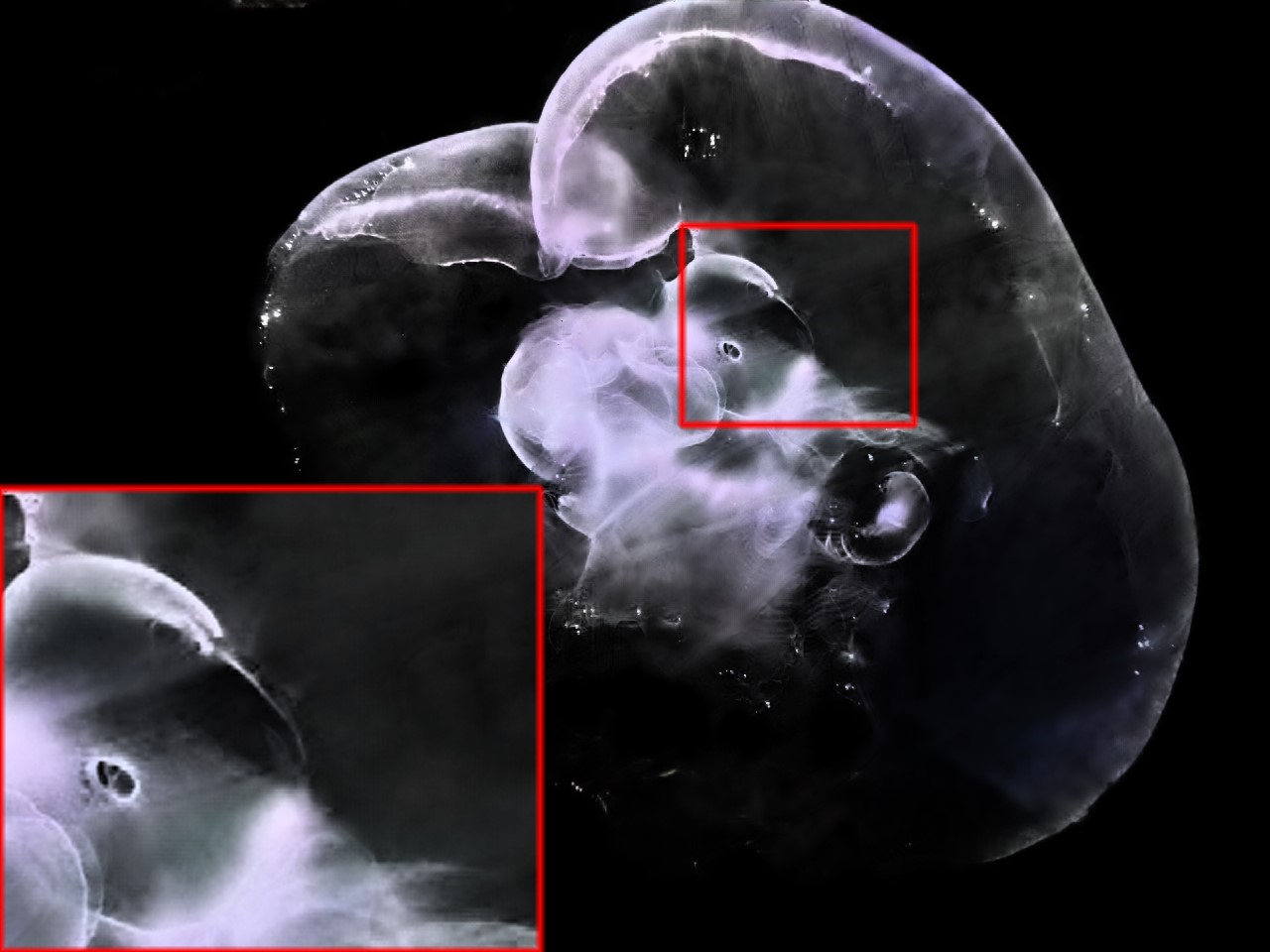} 
    \end{subfigure}
    \hfill
    \begin{subfigure}[t]{0.19\textwidth}
        \centering
        \includegraphics[width=\linewidth]{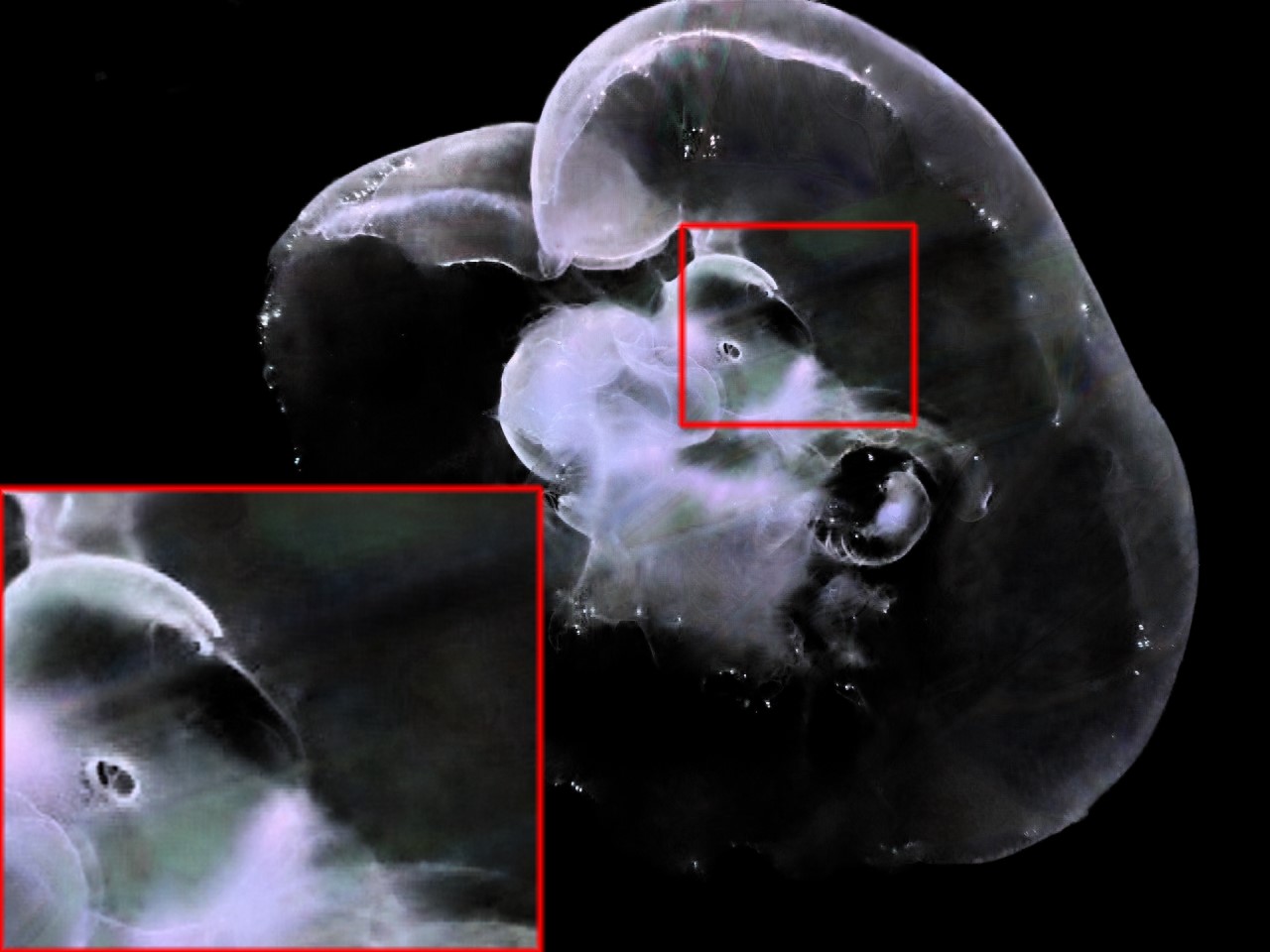} 
    \end{subfigure}
    \hfill
    \begin{subfigure}[t]{0.19\textwidth}
        \centering
        \includegraphics[width=\linewidth]{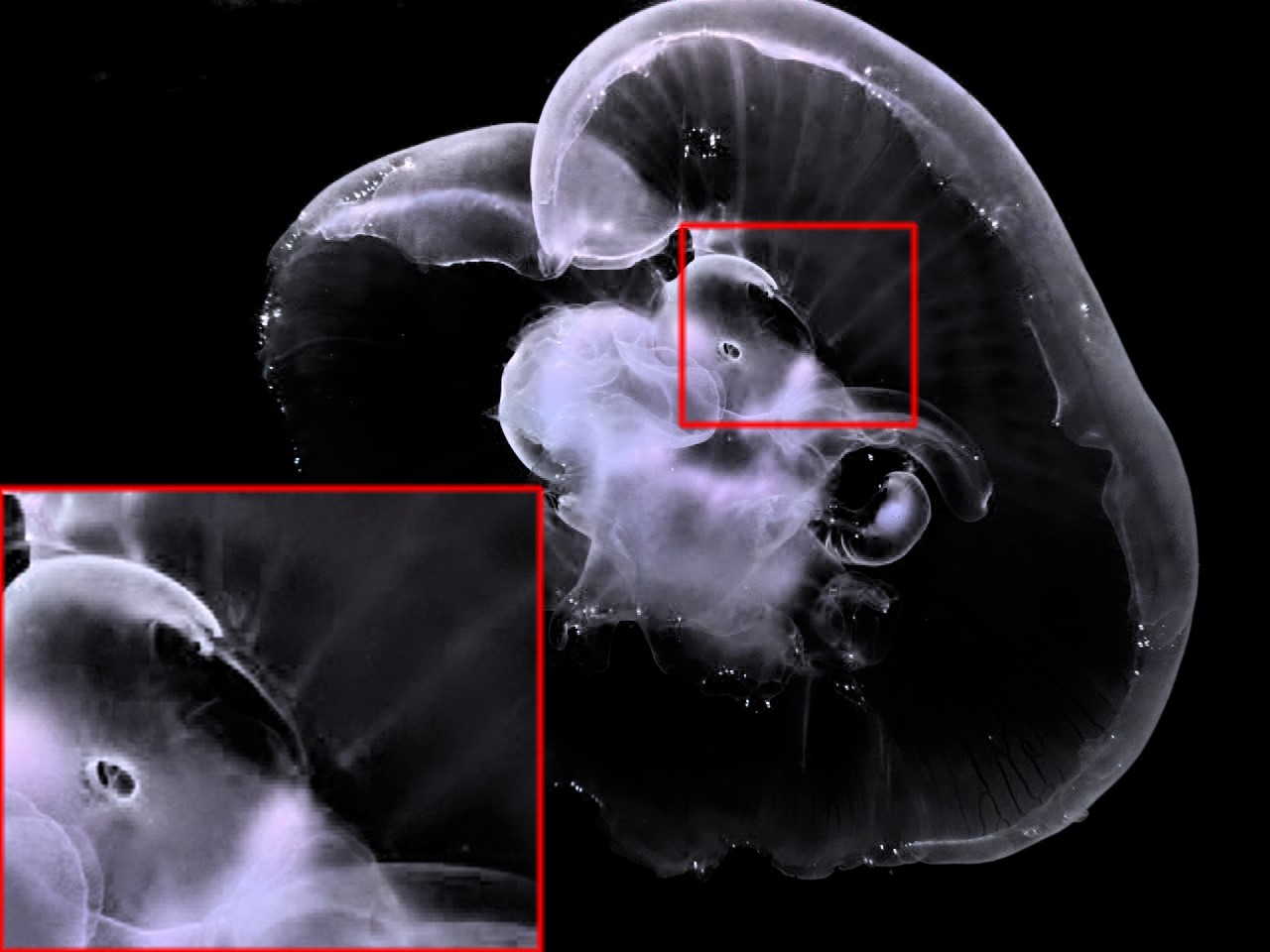} 
    \end{subfigure}
    \hfill
    \vspace{0.2cm}
    
    
    
    \begin{subfigure}[t]{0.19\textwidth}
        \centering
        \includegraphics[width=\linewidth]{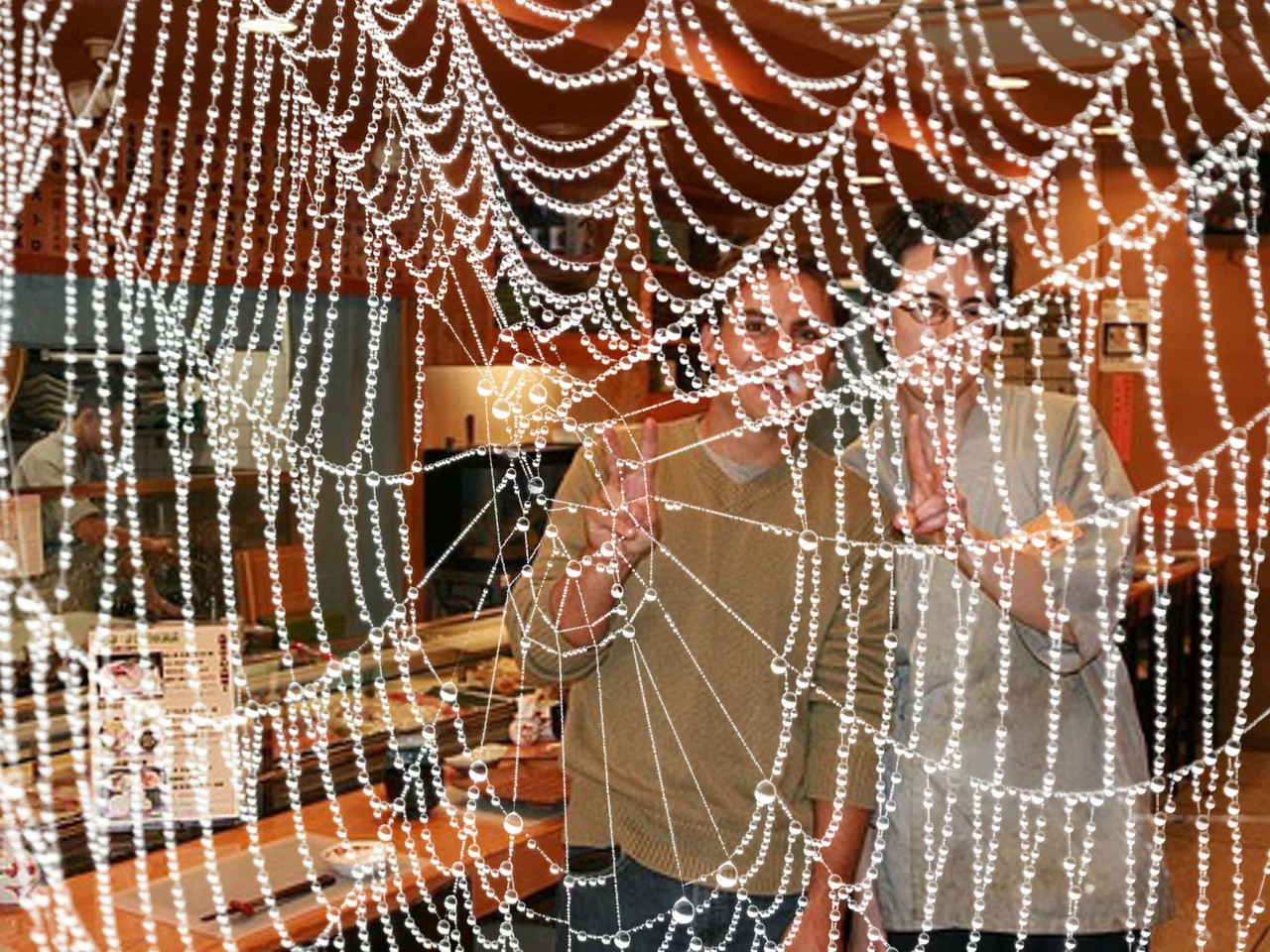} 
    \end{subfigure}
    \hfill
    \begin{subfigure}[t]{0.19\textwidth}
        \centering
        \includegraphics[width=\linewidth]{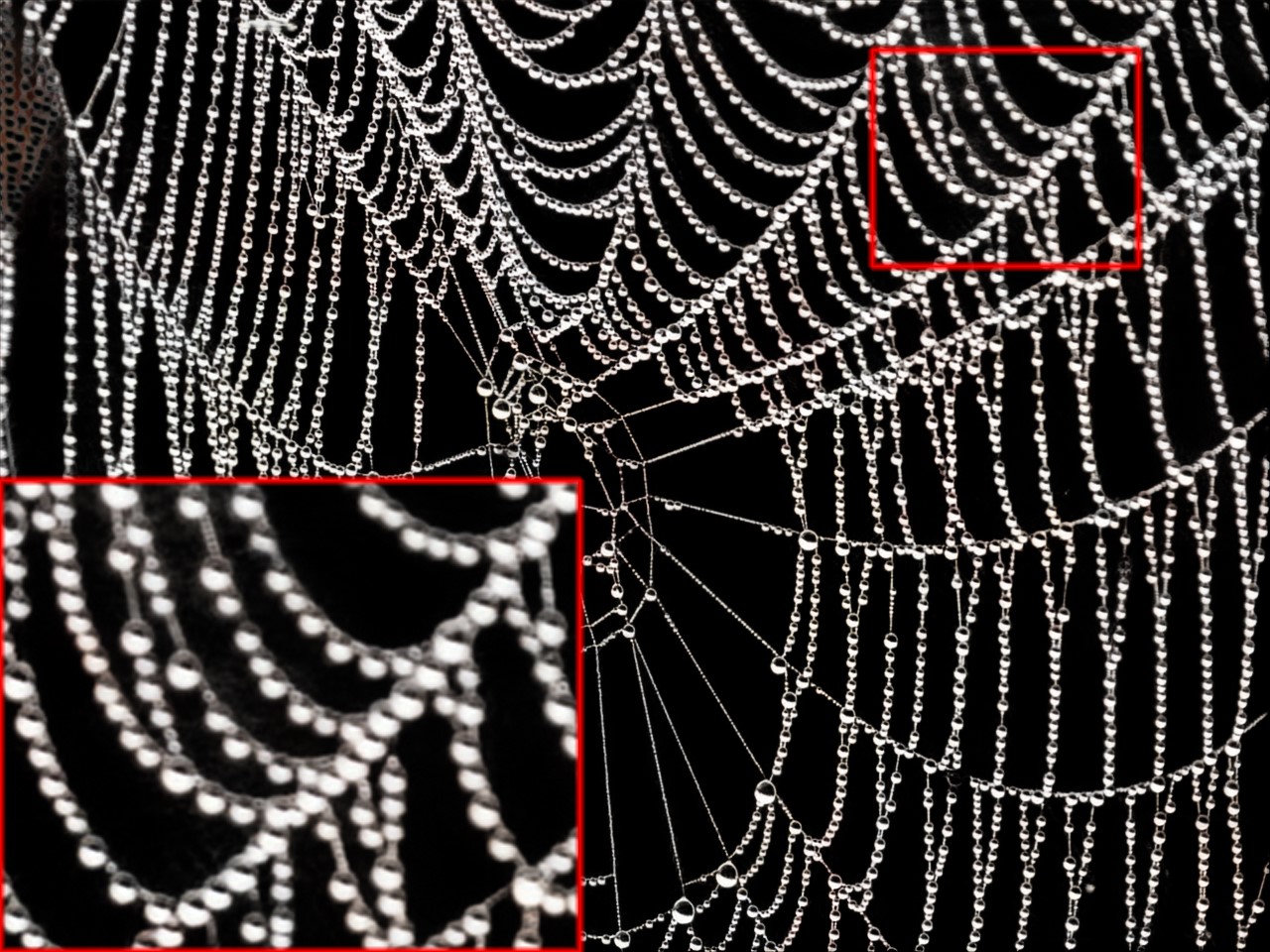} 
    \end{subfigure}
    \hfill
    \begin{subfigure}[t]{0.19\textwidth}
        \centering
        \includegraphics[width=\linewidth]{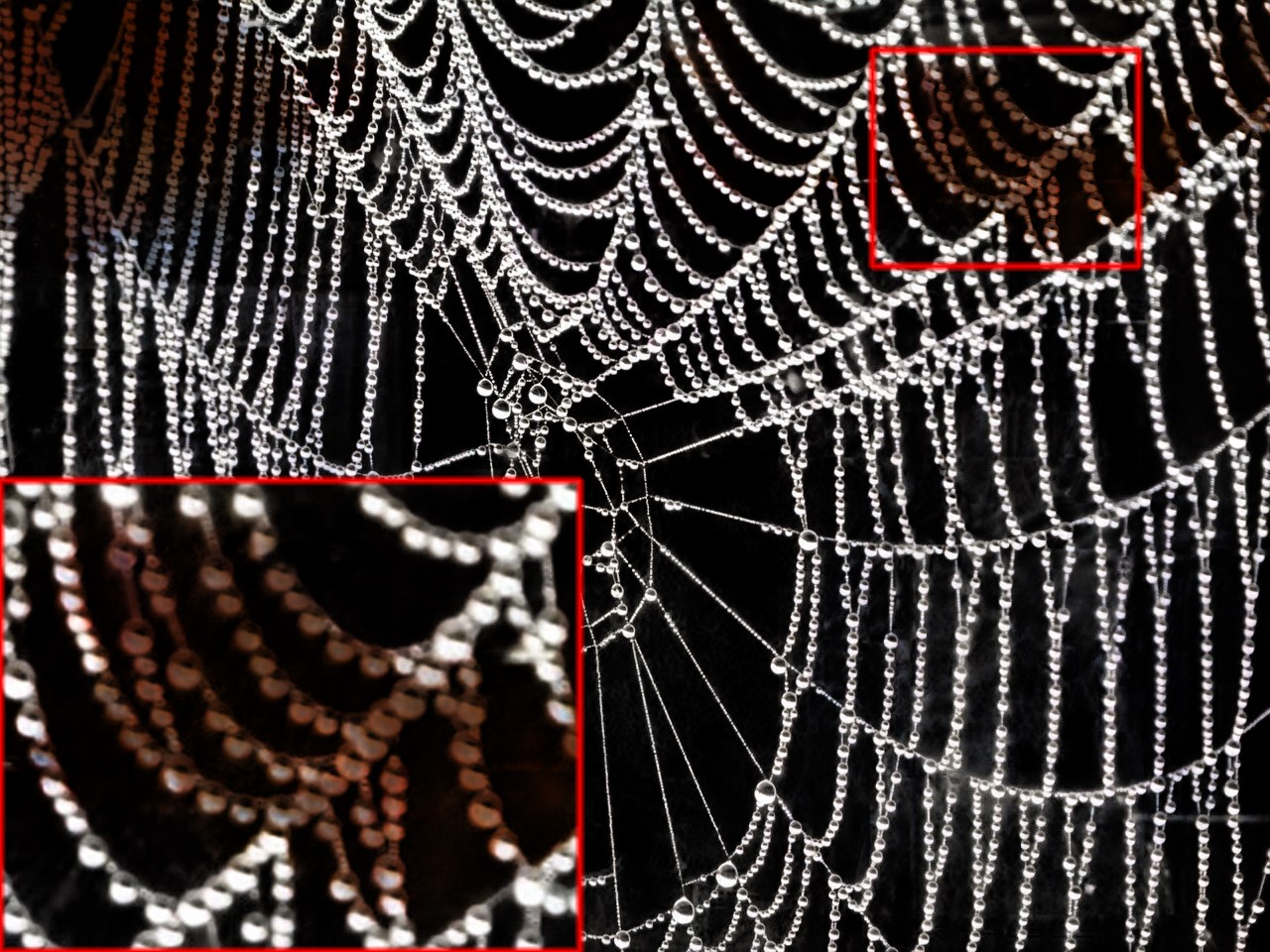} 
    \end{subfigure}
    \hfill
    \begin{subfigure}[t]{0.19\textwidth}
        \centering
        \includegraphics[width=\linewidth]{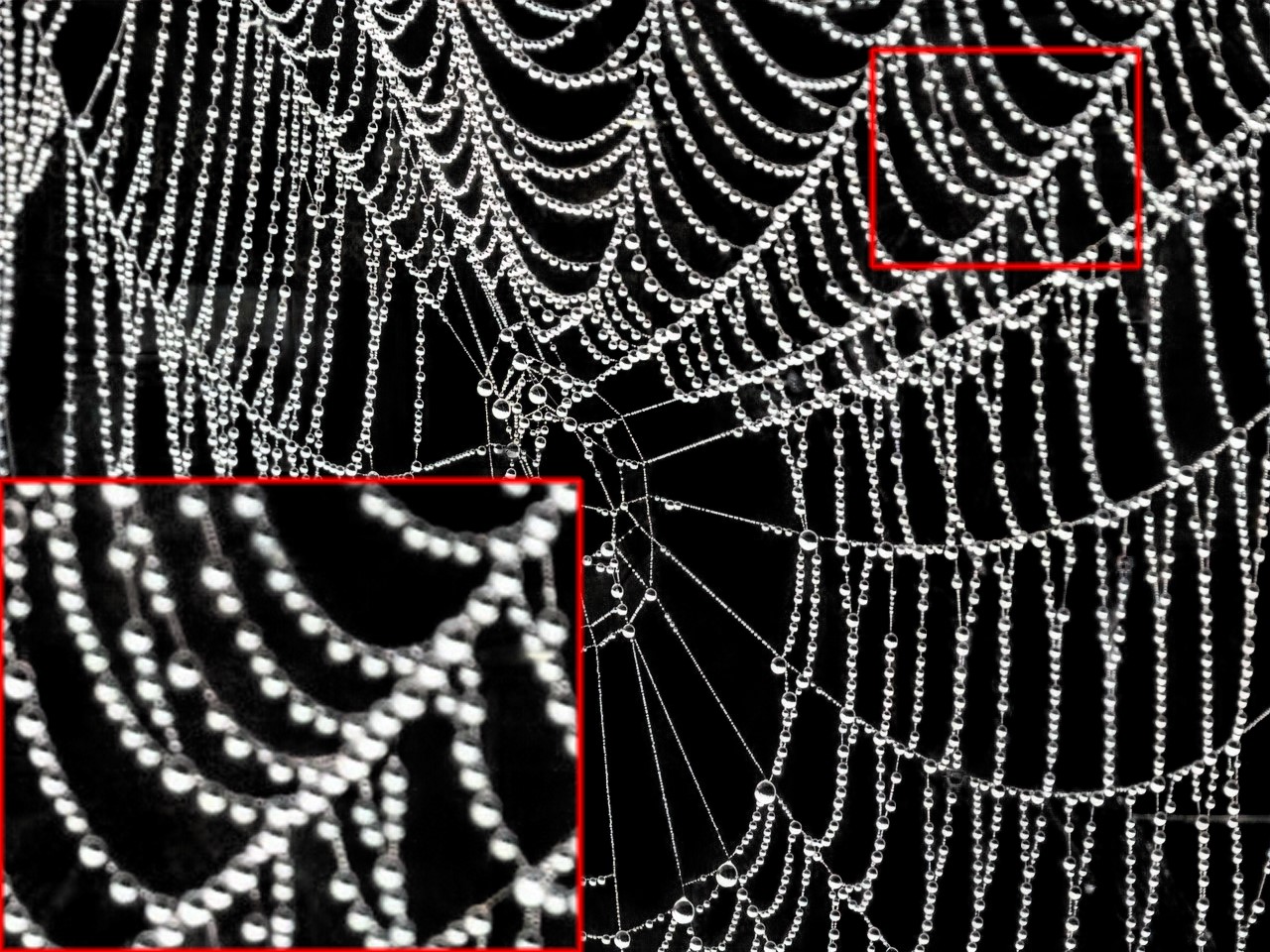} 
    \end{subfigure}
    \hfill
    \begin{subfigure}[t]{0.19\textwidth}
        \centering
        \includegraphics[width=\linewidth]{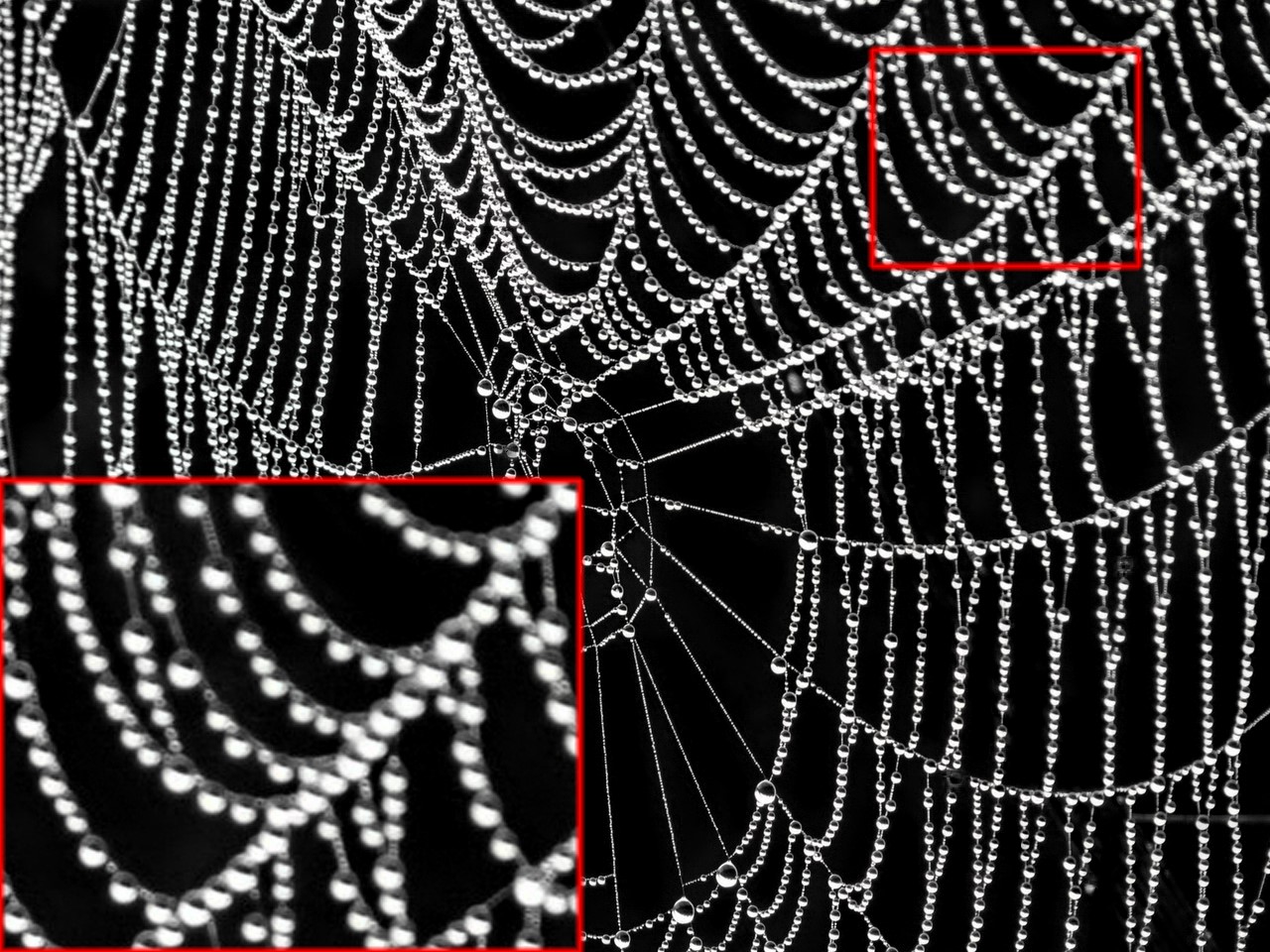} 
    \end{subfigure}
    \hfill
    \vspace{0.2cm}
    
    \begin{subfigure}[t]{0.19\textwidth}
        \centering
        \includegraphics[width=\linewidth]{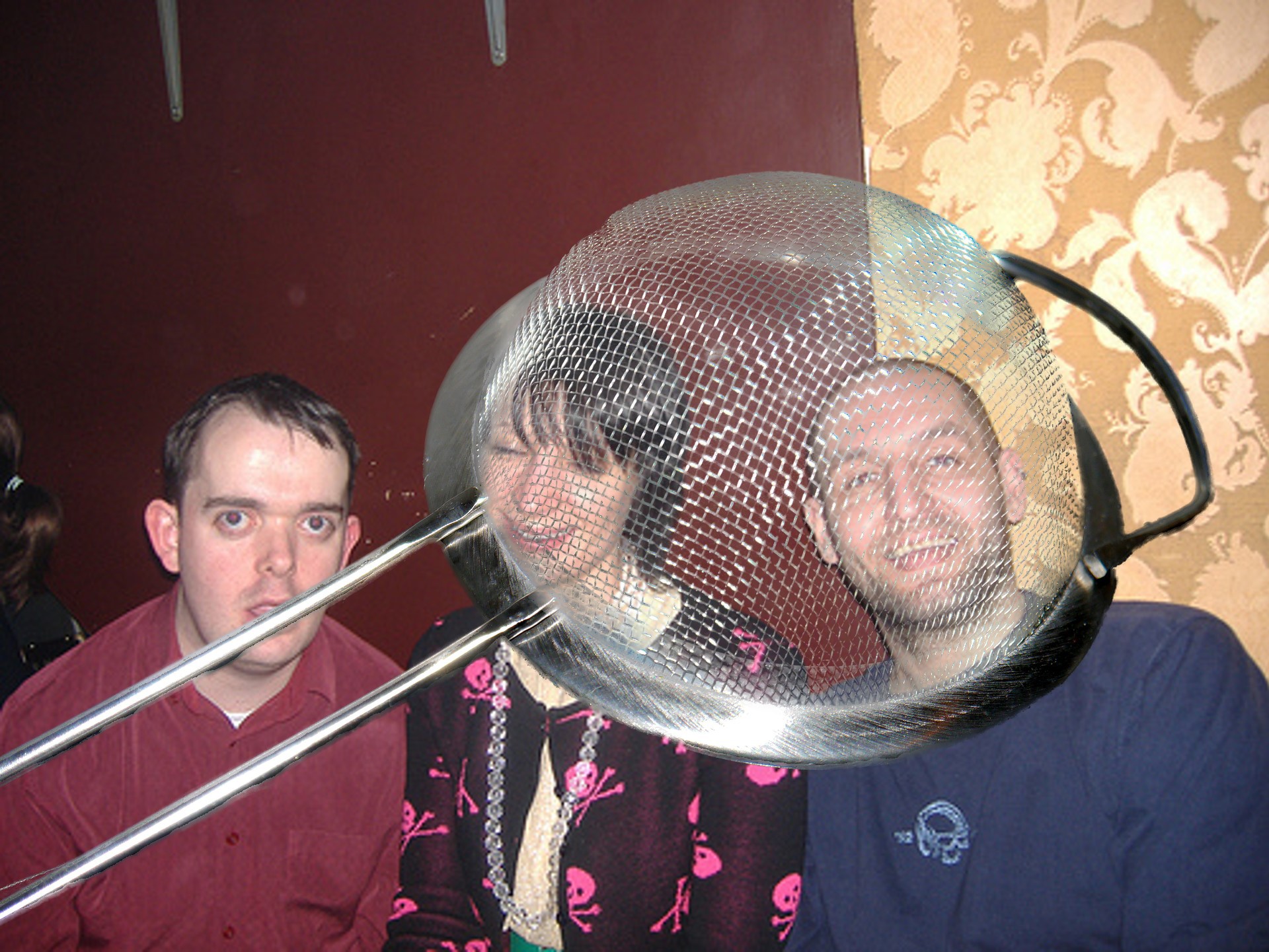} 
    \end{subfigure}
    \hfill
    \begin{subfigure}[t]{0.19\textwidth}
        \centering
        \includegraphics[width=\linewidth]{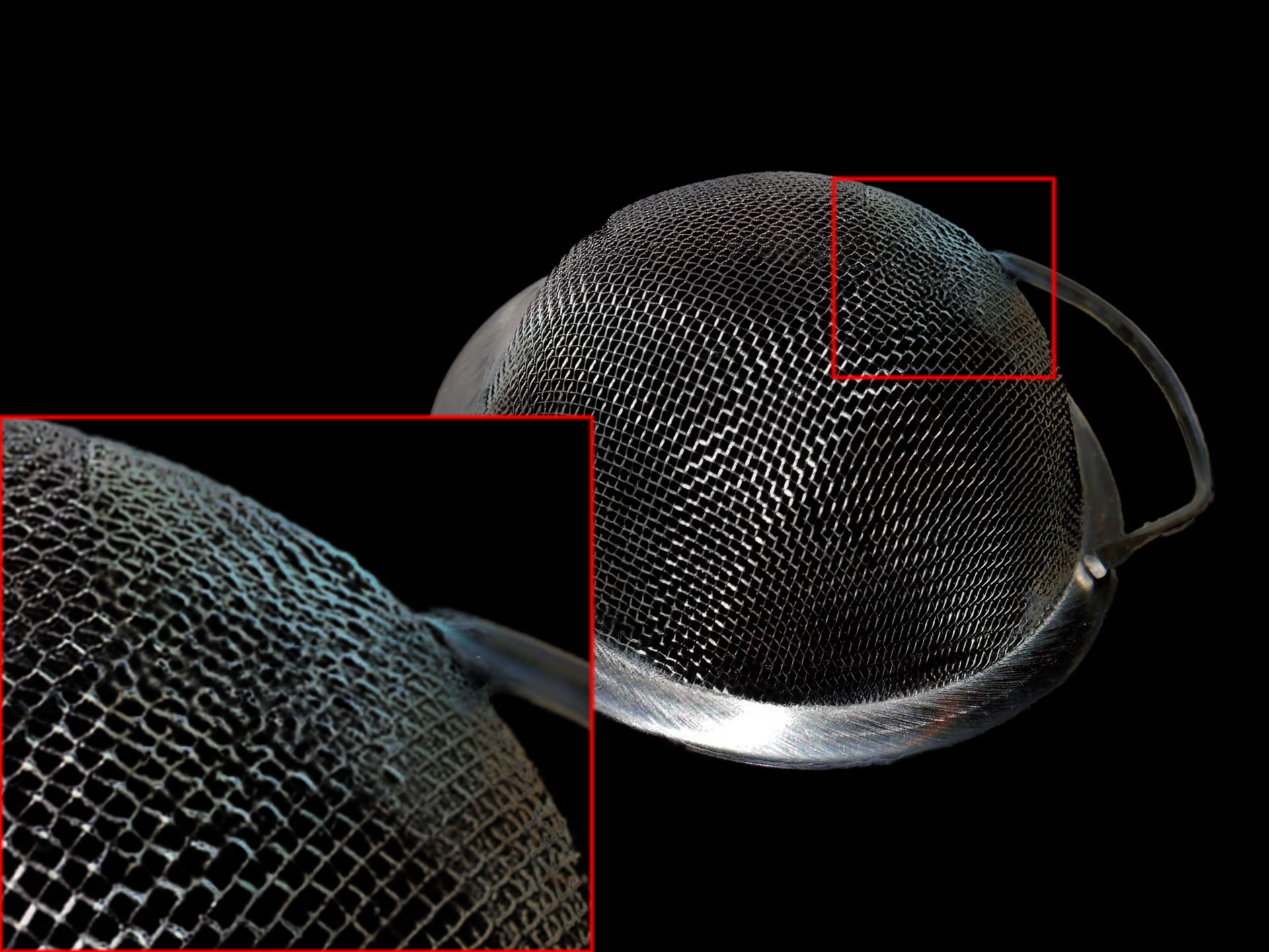} 
    \end{subfigure}
    \hfill
    \begin{subfigure}[t]{0.19\textwidth}
        \centering
        \includegraphics[width=\linewidth]{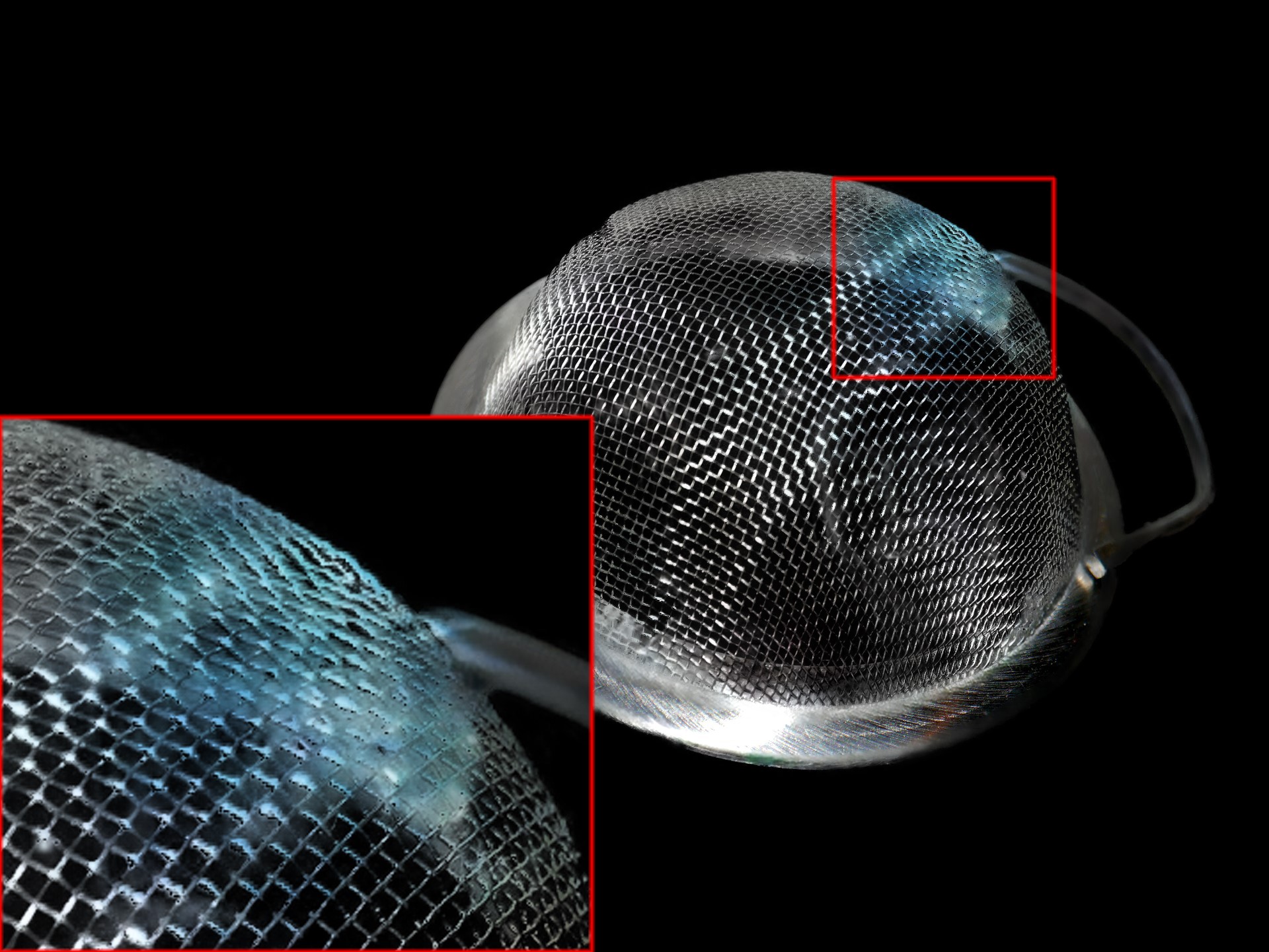} 
    \end{subfigure}
    \hfill
    \begin{subfigure}[t]{0.19\textwidth}
        \centering
        \includegraphics[width=\linewidth]{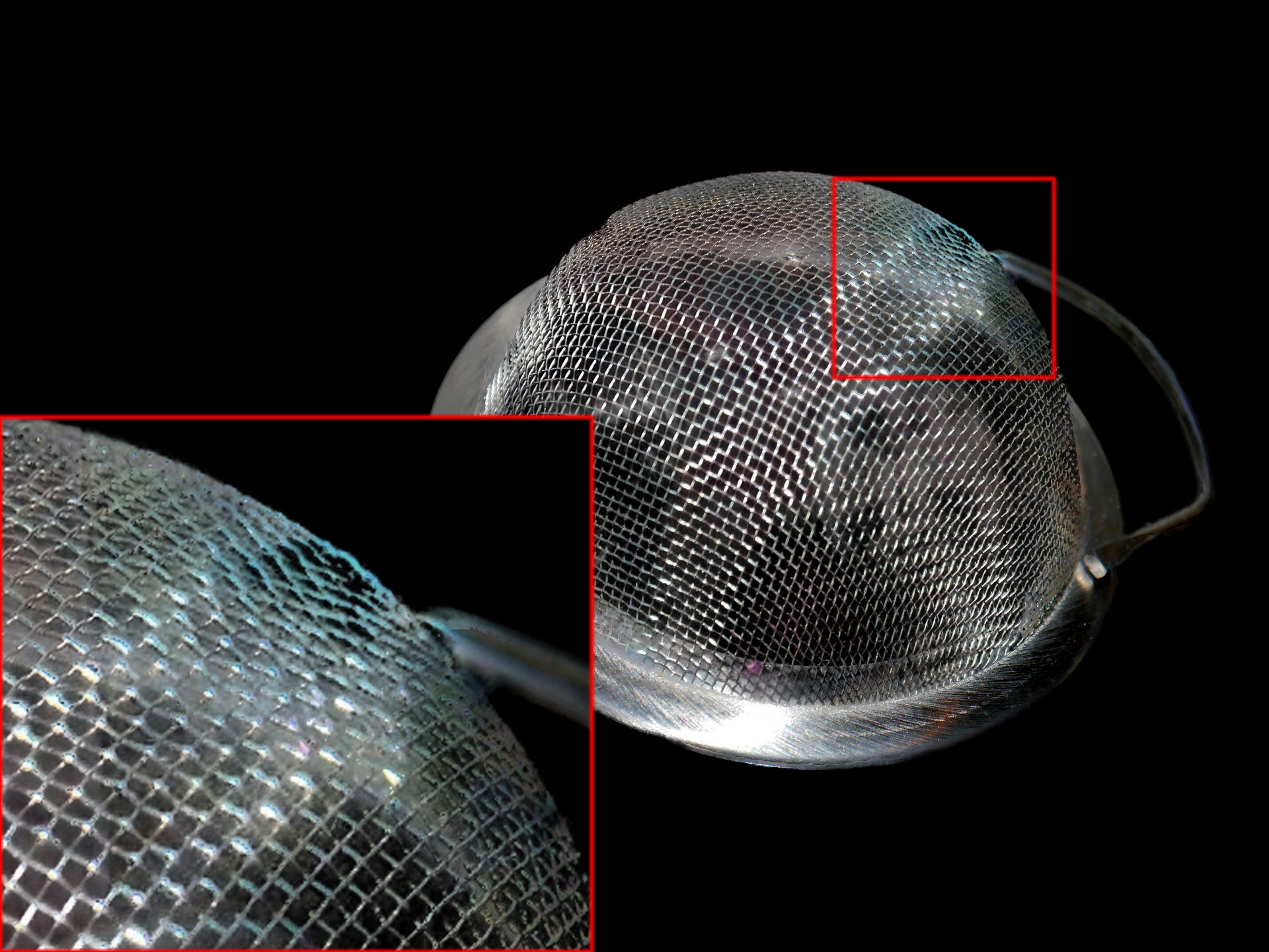} 
    \end{subfigure}
    \hfill
    \begin{subfigure}[t]{0.19\textwidth}
        \centering
        \includegraphics[width=\linewidth]{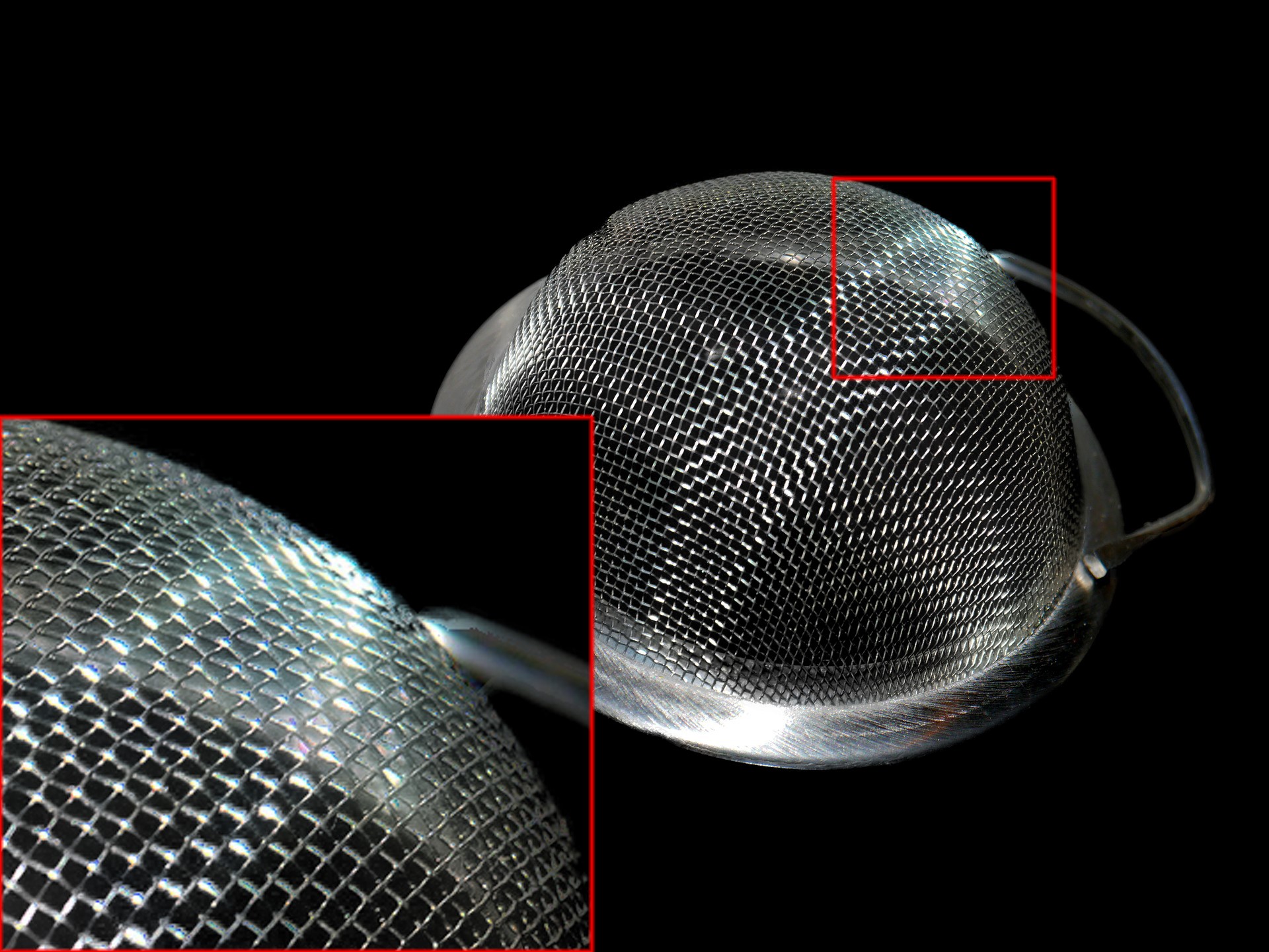} 
    \end{subfigure}
    \hfill
    \vspace{0.2cm}
    
    \begin{subfigure}[t]{0.19\textwidth}
        \centering
        \includegraphics[width=\linewidth]{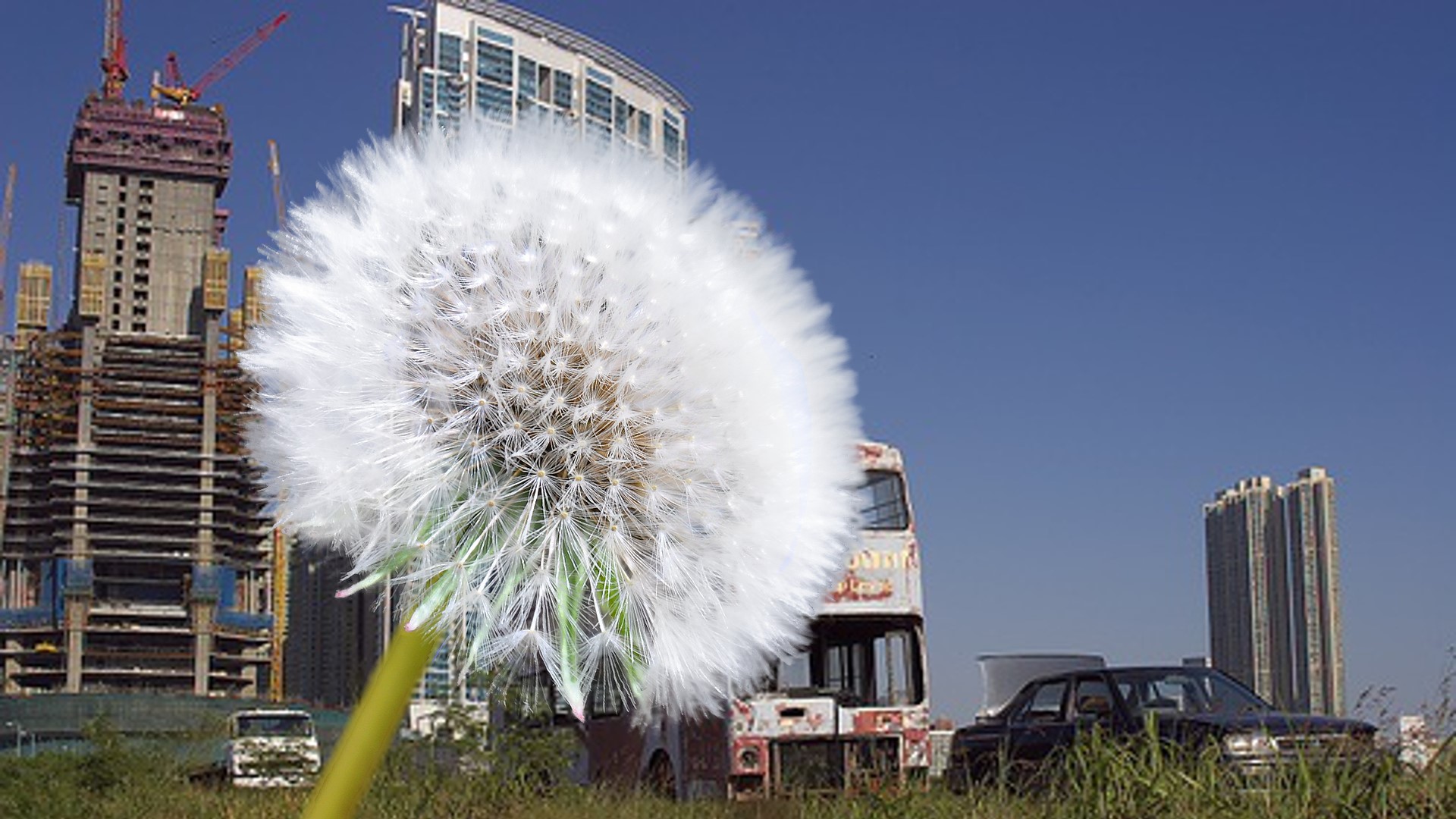} 
    \end{subfigure}
    \hfill
    \begin{subfigure}[t]{0.19\textwidth}
        \centering
        \includegraphics[width=\linewidth]{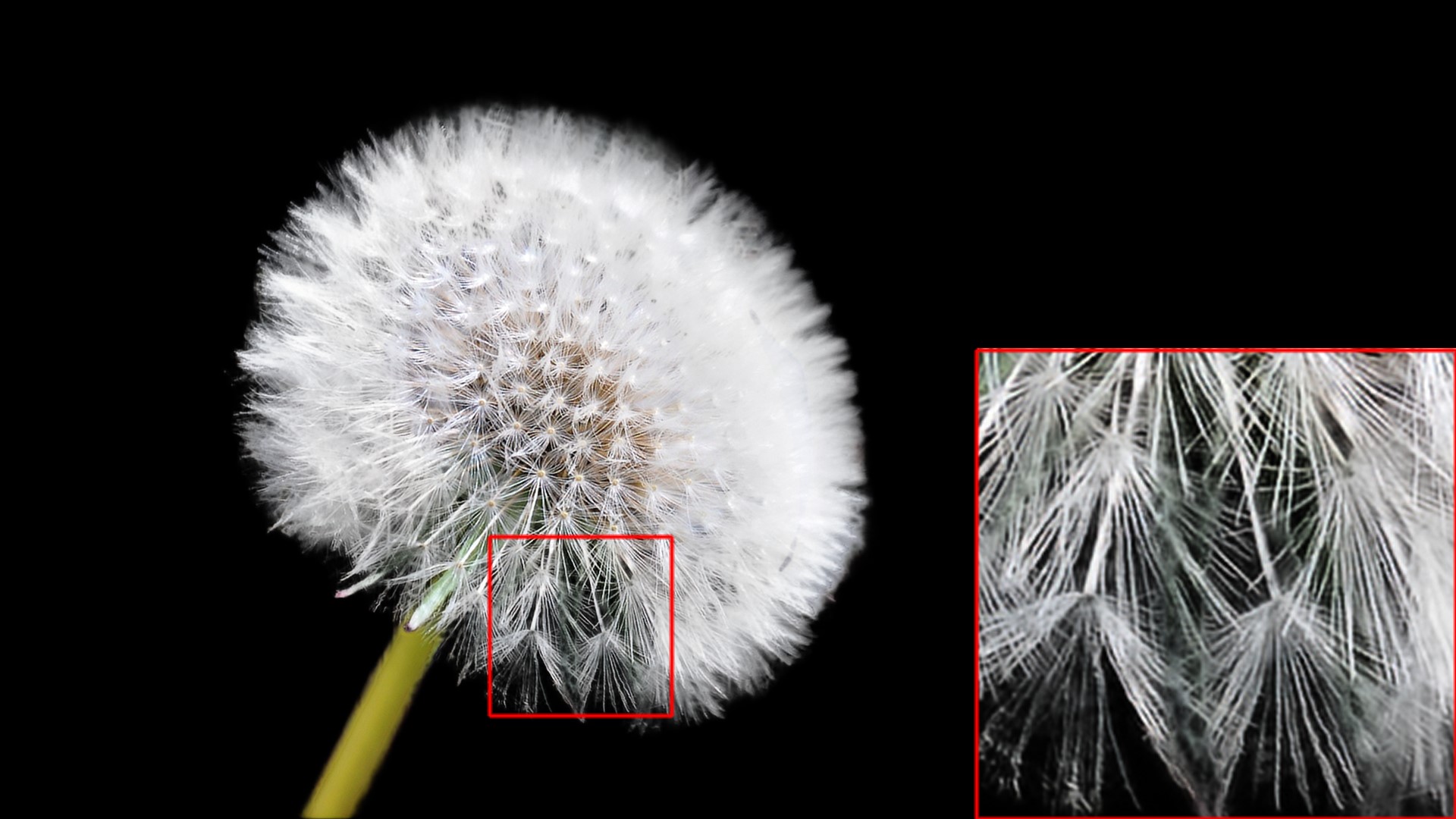} 
    \end{subfigure}
    \hfill
    \begin{subfigure}[t]{0.19\textwidth}
        \centering
        \includegraphics[width=\linewidth]{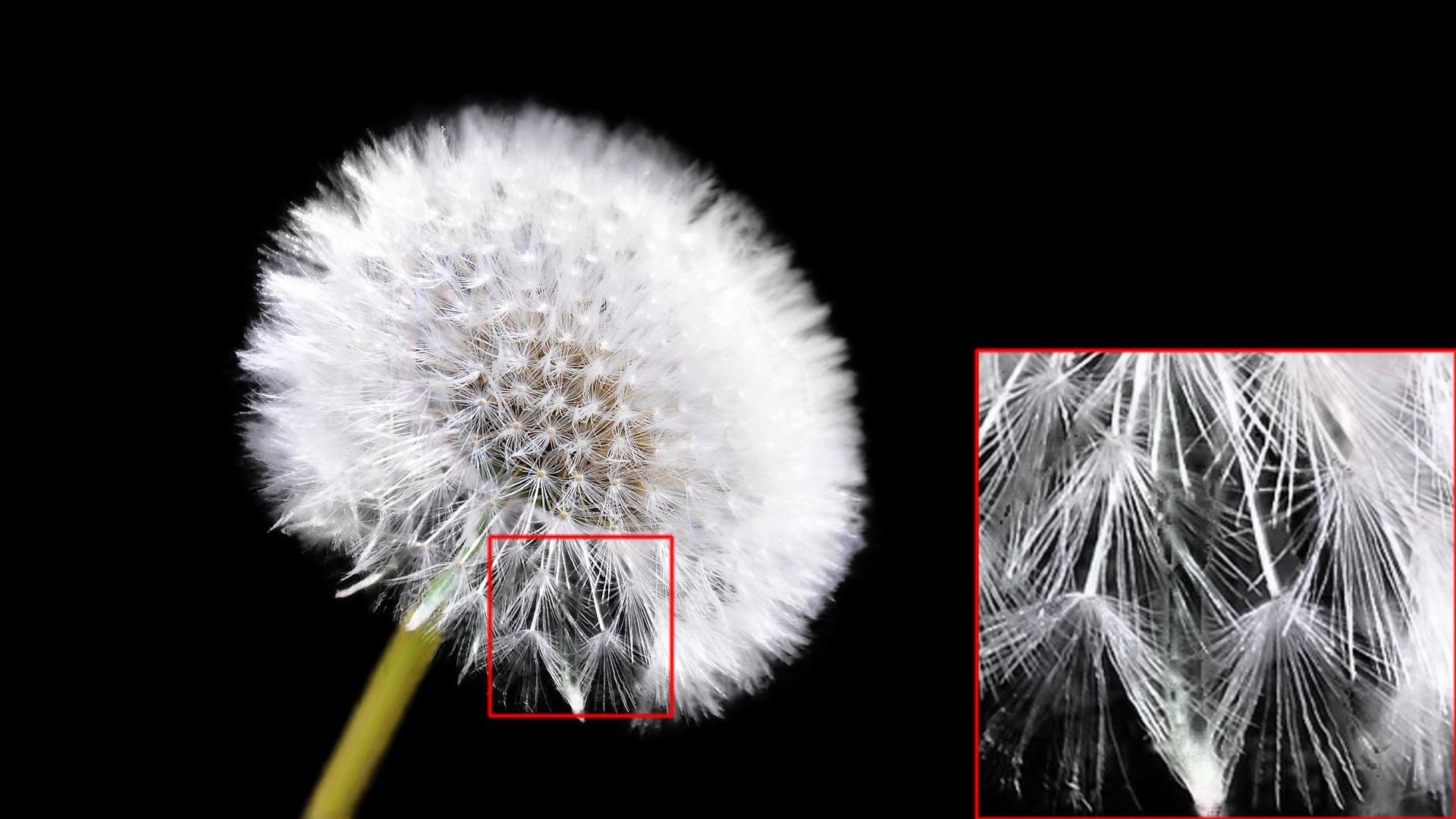} 
    \end{subfigure}
    \hfill
    \begin{subfigure}[t]{0.19\textwidth}
        \centering
        \includegraphics[width=\linewidth]{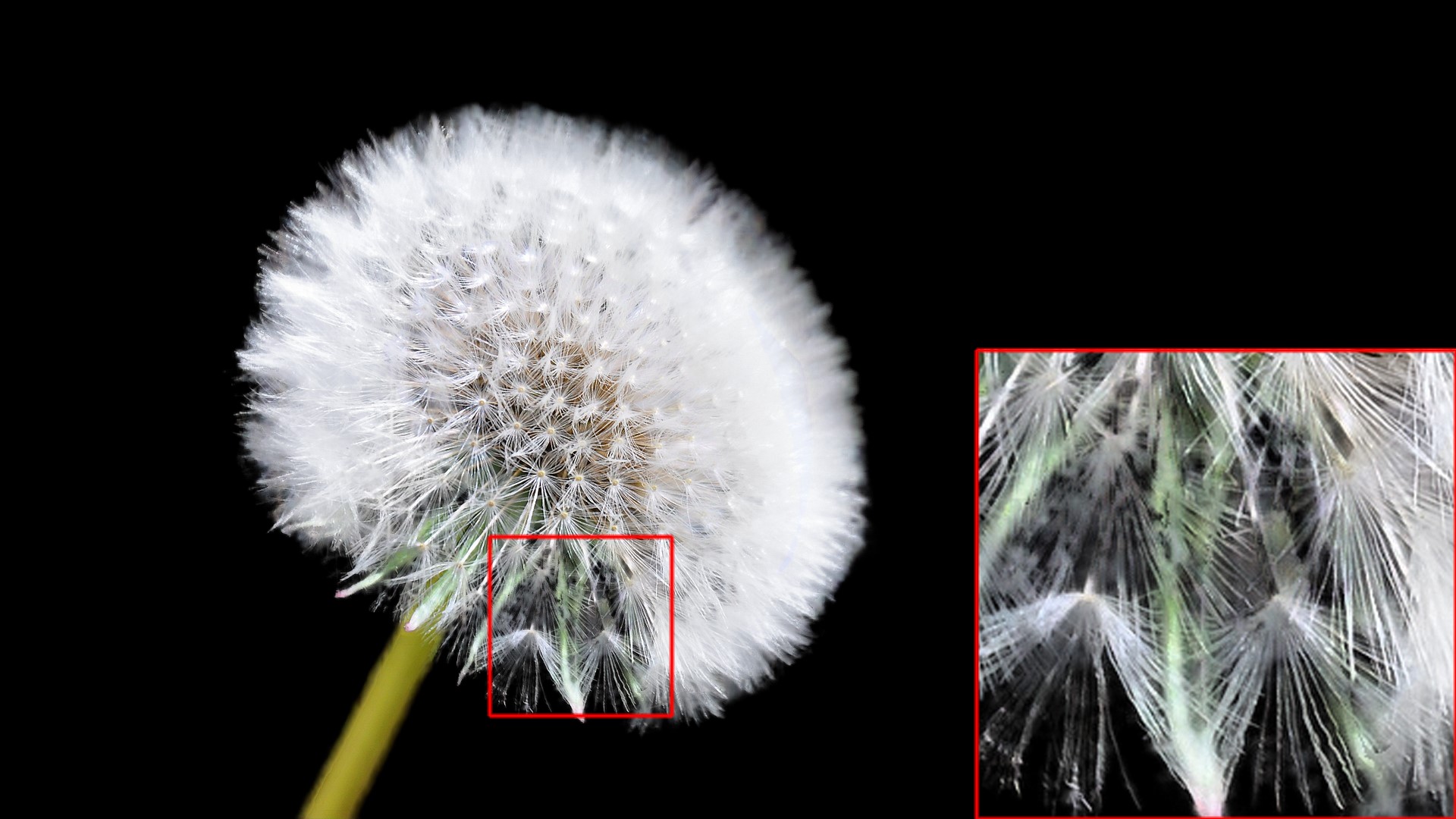} 
    \end{subfigure}
    \hfill
    \begin{subfigure}[t]{0.19\textwidth}
        \centering
        \includegraphics[width=\linewidth]{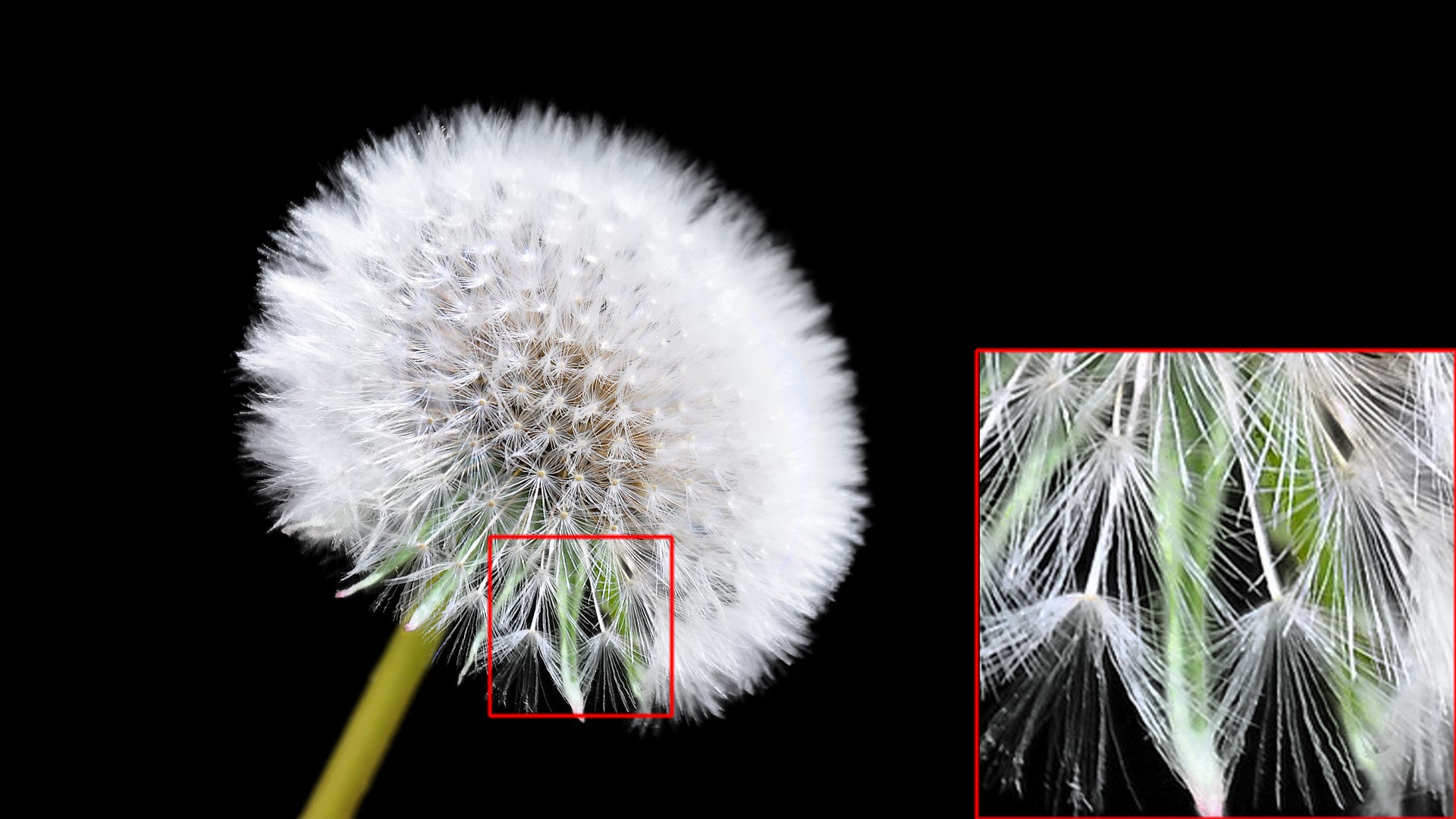} 
    \end{subfigure}
    \hfill
    \vspace{0.2cm}
    
    \begin{subfigure}[t]{0.19\textwidth}
        \centering
        \includegraphics[width=\linewidth]{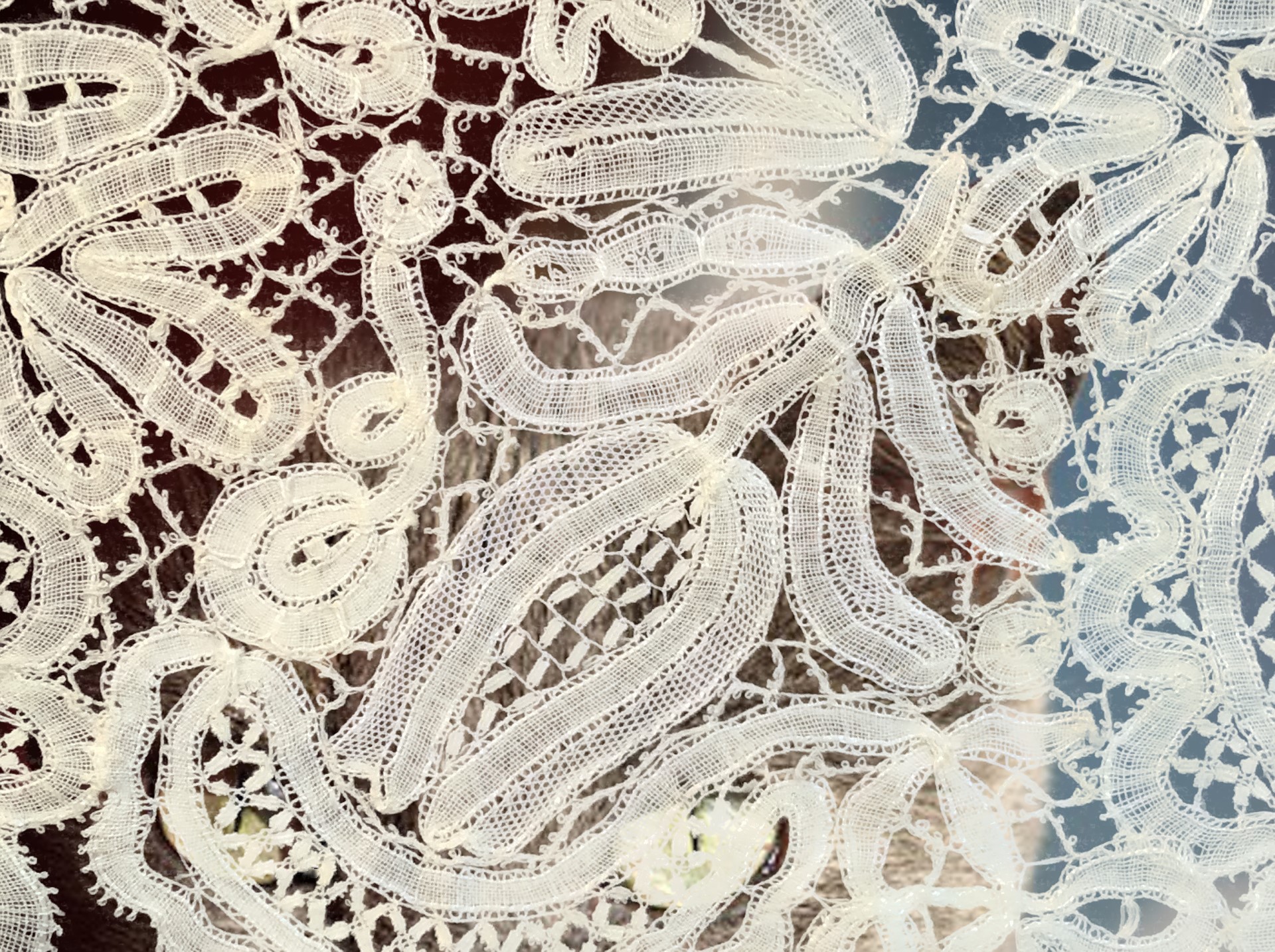} 
    \end{subfigure}
    \hfill
    \begin{subfigure}[t]{0.19\textwidth}
        \centering
        \includegraphics[width=\linewidth]{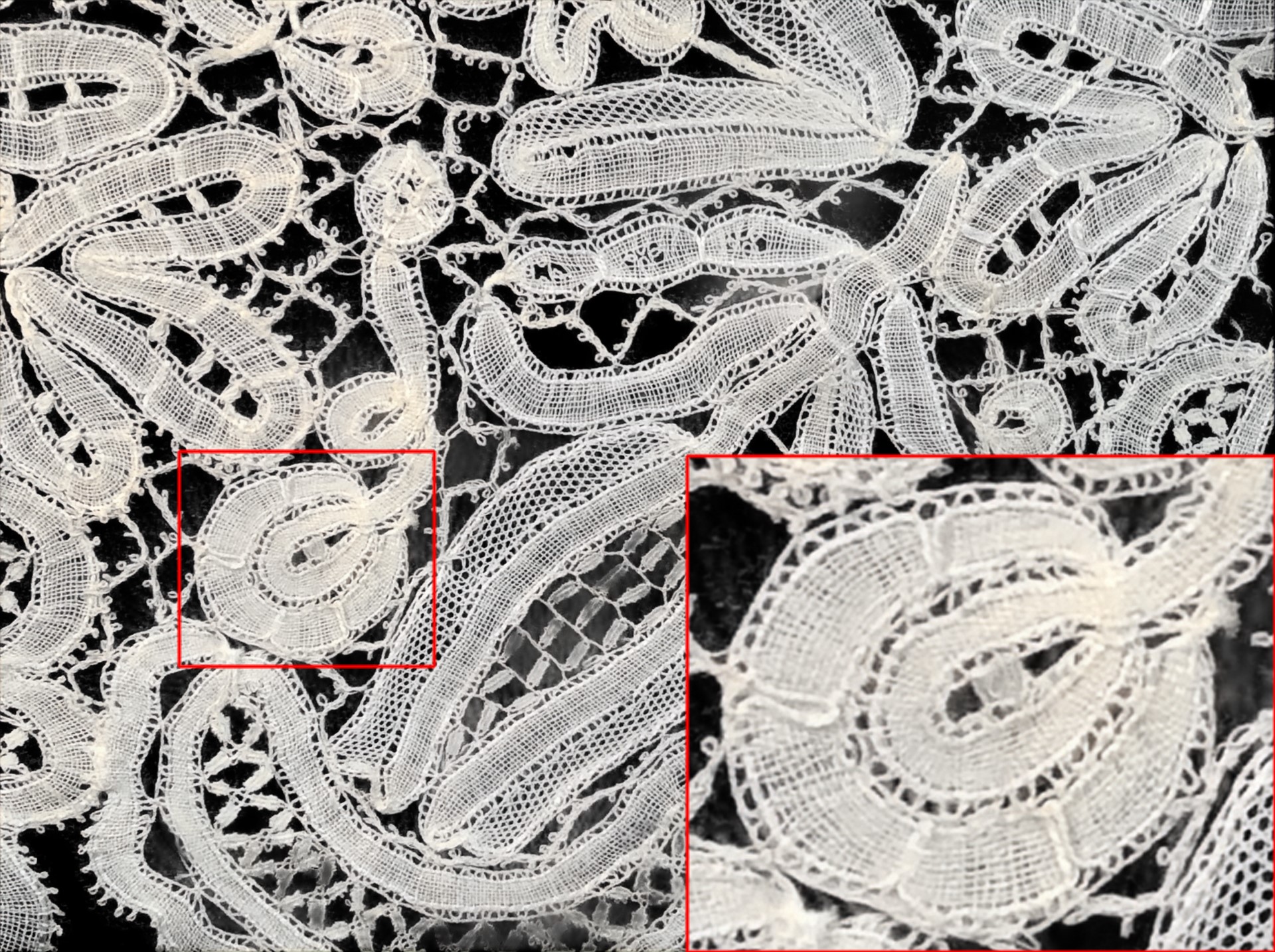} 
    \end{subfigure}
    \hfill
    \begin{subfigure}[t]{0.19\textwidth}
        \centering
        \includegraphics[width=\linewidth]{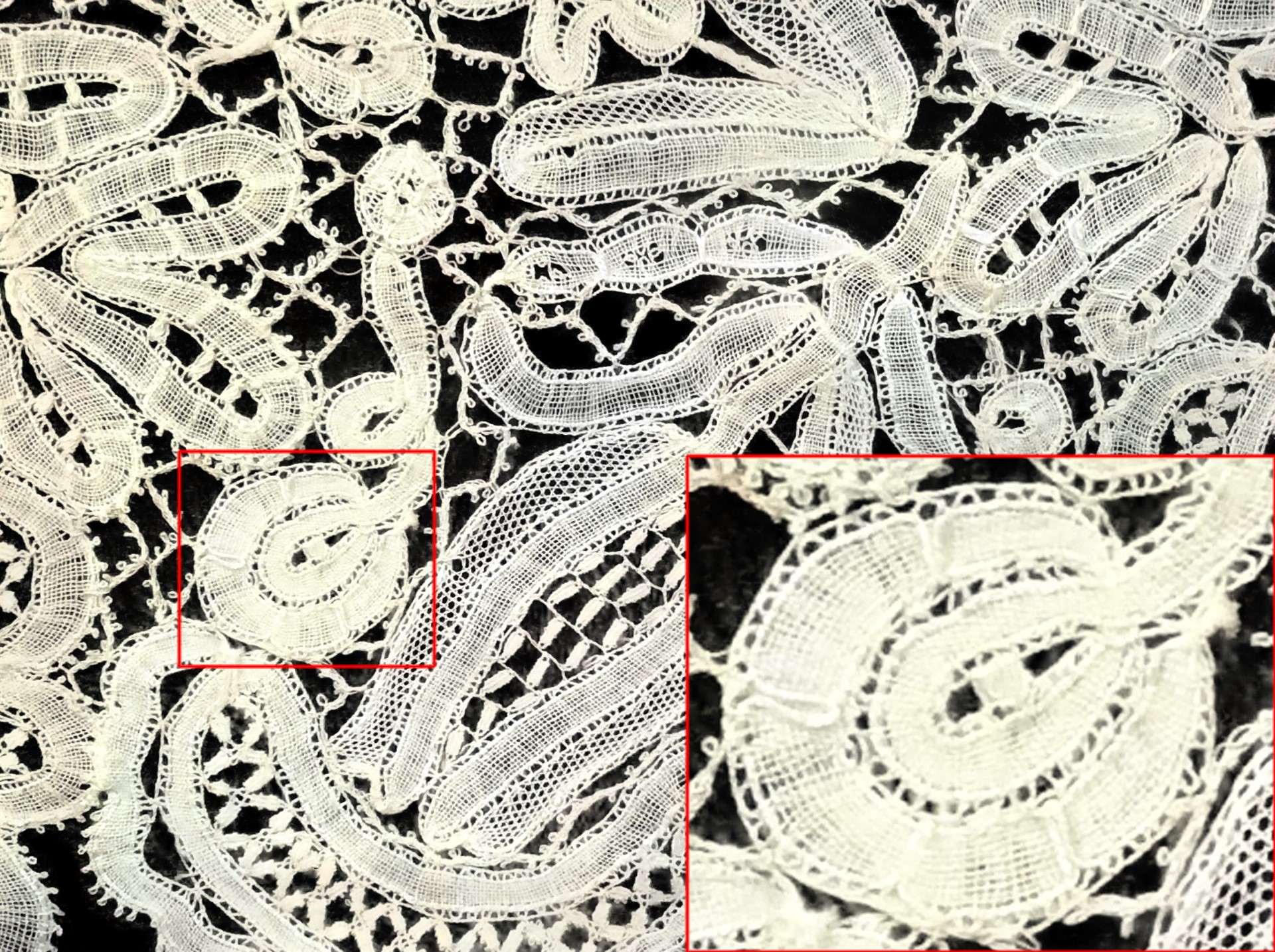} 
    \end{subfigure}
    \hfill
    \begin{subfigure}[t]{0.19\textwidth}
        \centering
        \includegraphics[width=\linewidth]{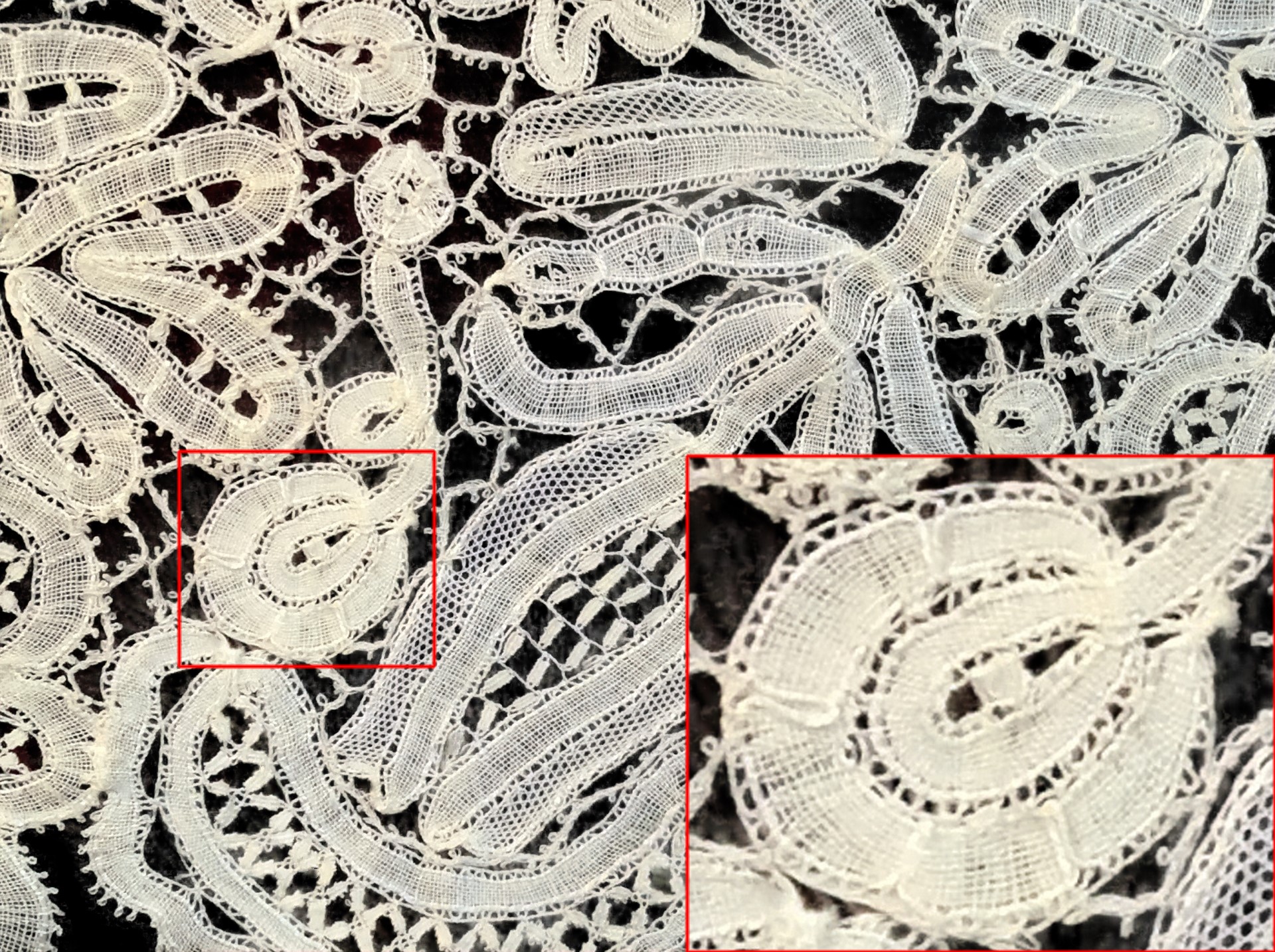} 
    \end{subfigure}
    \hfill
    \begin{subfigure}[t]{0.19\textwidth}
        \centering
        \includegraphics[width=\linewidth]{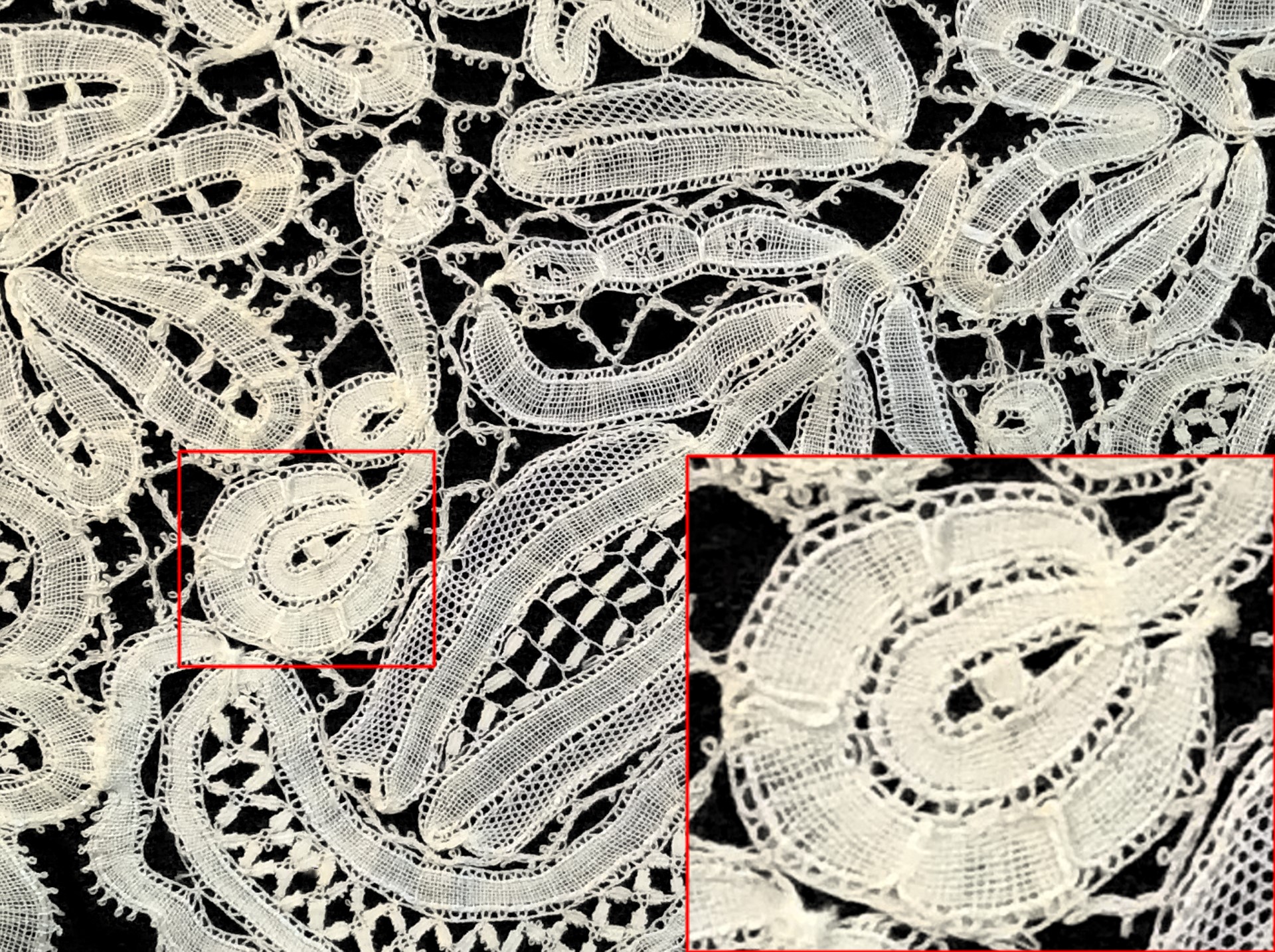} 
    \end{subfigure}
    \hfill
    \vspace{0.2cm}

    \caption{Visual comparison on the Composition-1k dataset. From left to right: Input image, Compositions from Context-Aware \cite{DBLP:conf/iccv/Hou019}, Samplenet \cite{DBLP:conf/cvpr/TangAOGA19}, Ours, Ground-truth. 
    }
    \label{fig:results}
\end{figure*}

We evaluate our proposed method on the Composition-1k dataset \cite{DBLP:conf/cvpr/XuPCH17}, with regards to the alpha matte prediction, as well as the foreground and background color prediction. The Composition-1k dataset consists of $50$ unique foreground images that have been composited into new images using $20$ predefined backgrounds each, resulting in $1000$ testing images in total. To generate our results, we use several different published methods for alpha prediction and use them as initial guess for our method. Using these predictions, we then generate foreground colors, background colors and alphas over $5$ iterations in total, which we experimentally found to lead to good convergence while keeping computation time low.

\subsection{Foreground and background color prediction}
Due to the composited nature of the Composition-1k dataset, we have the ground-truth background and foreground colors available. Therefore, we can calculate metrics for prediction on this dataset. As metrics, we choose the MSE and SAD of $\alpha \times F$ and $(1 - \alpha) \times B$ as introduced in \cite{DBLP:conf/cvpr/PriceMC10}. However, due to the nature of the dataset, which can contain wrong foreground colors in all pixels where the alpha is $0$, we multiply by the ground-truth alpha. This also disentangles the performance of all evaluated methods from their performance in predicting the alpha. We further only consider pixels in the unknown trimap area.\\
We compare our results against Closed-form Matting \cite{DBLP:conf/cvpr/LevinLW06}, Context-Aware Matting \cite{DBLP:conf/iccv/Hou019} and Samplenet \cite{DBLP:conf/cvpr/TangAOGA19}. Closed-form Matting can predict the alpha matte by solving a sparse linear system of equations. As a second step and given an alpha as input, it is possible to solve this equation for the foreground and background colors instead of the alpha. For fair comparisons, we use the superior alpha predictions from GCA-Matting \cite{DBLP:journals/corr/abs-2001-04069} as input. Context-Aware Matting predicts the foreground and alpha simultaneously through a deep neural network. By contrast, Samplenet does a full background, foreground and alpha prediction sequentially using 3 networks. Additionally, we can use the input image as a baseline comparison. We use the same initial alpha predictions for our method as in the evaluation of the alpha prediction, as are shown in table \ref{tab:adobe_color_results}.\\
As can be seen in the table, we achieve the overall best results for the color prediction by a wide margin, especially in the prediction of the foreground colors. The background predictions of Samplenet are marginally better than ours according to the SAD, but it performs worse in foreground prediction.
We can also observe that the quality of the initial alpha prediction has a large impact on the resulting quality of the foreground and background predictions. The better the initial alpha prediction is, the better our method performs. However, even alpha predictions from the outdated KNN-Matting lead to foreground colors that are almost as good as the colors naively taken from the input image and noticeable better background colors. We also outperform the Closed-form solution for the foreground and background colors, even when given the GCA-Matting alpha prediction as input. Furthermore, the Closed-form solution is quite slow in comparison to modern deep learning methods, which is non-optimal for any interactive application. Visual comparisons can be seen in figure \ref{fig:results} and in the supplementary materials.

\begin{table}[t]
    \centering
    \resizebox{.45\textwidth}{!}{
    \begin{tabular}{ lcccc } 
    \hline
     & \multicolumn{2}{c}{Foreground} & \multicolumn{2}{c}{Background}\\
     \hline
     Methods & SAD & MSE ($10^4$) & SAD & MSE ($10^4$) \\\hline
     Input Image & 58.32 & 26.49 & 57.90 & 26.12 \\
     KNN \cite{DBLP:conf/cvpr/ChenLT12} + Ours  & 59.91 & 49.25 & 36.77 & 15.19\\
     AlphaGAN \cite{DBLP:conf/bmvc/LutzAS18} + Ours  & 44.27 & 27.90 & 40.12 & 16.93\\
     IF \cite{DBLP:conf/cvpr/AksoyAP17} + Ours  & 37.93 & 31.98 & 29.49 & 15.82\\
     GCA \cite{DBLP:journals/corr/abs-2001-04069} + CF \cite{DBLP:conf/cvpr/LevinLW06} & 31.98 & 23.15 & 29.40 & 10.69\\
     Context-Aware \cite{DBLP:conf/iccv/Hou019} & 46.93 & 18.02 & - & -\\
     SampleNet \cite{DBLP:conf/cvpr/TangAOGA19} & 42.68 & 29.26 & \textbf{24.59} & 7.99\\
     GCA \cite{DBLP:journals/corr/abs-2001-04069} + Ours & \textbf{28.32} & \textbf{12.10} & 25.07 & \textbf{5.97}\\
     \hline
    \end{tabular}
    }
    \caption{Quantitative results of the foreground and background prediction on the Composition-1k dataset. Best results are emphasized in bold. Note that not all images could be predicted for KNN Matting and Information-flow Matting due to trimaps incompatible with these methods.}
    \label{tab:adobe_color_results}
\end{table}

\subsection{Color prediction over several iterations}
Our method is a recurrent neural network and predicts update steps to the previous solution over several iterations to output new solutions. We compare our results for the color predictions in table \ref{tab:color_iterations}. As can be seen, our predictions get consecutively better over iterations until they saturate after $t = 5$.

\subsection{Manual editing}
To show the impact of the manual editing process, we compare the fully automatic output of our method to results we get when making only small edits during the process. For this, we take 5 of the images of the Composition-1k dataset where the automatic prediction generates sub-par results and spend less than a minute each on manually improving the intermediate alpha predictions. As can be seen in table \ref{tab:manual_edit}, even small edits focusing on the foreground massively improve the color prediction. Naturally, these edits could be done on the alpha predictions of other methods as well, however, only our method interlinks the alpha and color predictions in a way to propagate the changes from the alpha to the foreground and background color predictions automatically. In other methods the edits to the alpha would have to be replicated for the color predictions as well, increasing the amount of work. Please refer to the supplementary materials for the images and edits.

\subsection{User study}
To further evaluate the quality of our work, we conduct a user study comparing the foreground color prediction of our method with Samplenet \cite{DBLP:conf/cvpr/TangAOGA19} and Context-Aware Matting \cite{DBLP:conf/iccv/Hou019}. Following the approach of \cite{DBLP:conf/iccv/Hou019}, we take all 31 images of the real-world image dataset from Xu et al. \cite{DBLP:conf/cvpr/XuPCH17} and use the predicted alpha and foreground colors to composite a new image with a plain background. To ensure any differences between matting results are only due to the color predictions, we use the predicted alpha from Samplenet as initial guess for our method in the comparisons with Samplenet and similar for our comparisons with Context-Aware Matting.\\
We recruited 20 participants for our user study for each of the comparisons. Each participant was submitted to a short training session where the results of two methods was shown and the differences explained. This was done to help people with no prior matting experience spot the subtle differences in results.\\
Each participant conducted $31$ trials corresponding to all the images of the real world dataset. In each trial the original image was shown at the top with the composited results of the methods at the bottom. The results were shown one at a time and the participant could use buttons or the arrow keys to switch the bottom image between the result images. The participants were asked to choose the result which they found more accurate and realistic.\\
We calculated the preference rate of the participants of our results and show the mean preference rate and standard deviation in table \ref{tab:study}. As can be seen, the majority of participants preferred our results to those of Samplenet. However, when comparing to Context-Aware Matting, the results show no preference of one over the other. Some examples of our study can be seen in the supplementary materials. As can be seen, the color differences between our results and Context-Aware Matting are very minor, which explains the responses. However, our method still significantly outperforms Context-Aware Matting numerically in the Composition-1k dataset and offers the option to further improve the results interactively.

\begin{table}[t]
    \centering
    \begin{tabular}{ lcccc } 
    \hline
     & \multicolumn{2}{c}{Foreground} & \multicolumn{2}{c}{Background}\\
     \hline
     Iteration & SAD & MSE ($10^4$) & SAD & MSE ($10^4$)\\\hline
     1 & 42.06 & 16.34 & 35.17 & 10.67 \\
     2 & 31.18 & 12.87 & 27.57 & 7.20 \\
     3 & 29.18 & 12.24 & 25.83 & 6.33 \\
     4 & 28.41 & \textbf{12.05} & 25.24 & 6.05 \\
     5 & \textbf{28.32} & 12.10 & \textbf{25.07} & \textbf{5.97} \\
     \hline
    \end{tabular}
    \caption{Our results for foreground and background color prediction over iterations using the GCA alpha prediction as initial input. Best results are emphasized in bold.}
    \label{tab:color_iterations}
\end{table}

\subsection{Alpha matte prediction}
To show that our method does not degrade the quality of the alpha matte prediction of whichever method was used for the initial guess, we compare on the commonly used evaluation metrics \cite{DBLP:conf/cvpr/RhemannRWGKR09}. Our method slightly improves the quality of the alpha according to the metrics, but no to a significant amount. The table with the results of this evaluation can be found in the supplementary materials.


\subsection{Limitations}
The goal of this work is to introduce a lightweight method that can be used with any alpha prediction network to estimate the foreground and background colors that lead to compelling new composites. However, we do not refine the input alpha to a significant amount due to the small capacity of our network. In future work, it may be desirable to explore an updated network architecture that is able to further refine inadequate initial alpha predictions.\\
Further, as opposed to Samplenet, our method does not predict good background colors in areas where the background can not, or only barely, be seen in the image. This has no impact on new compositions and we do not claim to fully inpaint the background after the foreground has been removed. However, certain applications may find a fully inpainted background desirable, which we can not provide.

\begin{table}[t]
    \centering
    \begin{tabular}{ lcccc } 
    \hline
     & \multicolumn{2}{c}{Foreground} & \multicolumn{2}{c}{Background}\\
     \hline
      & SAD & MSE ($10^4$) & SAD & MSE ($10^4$)\\\hline
     Pre-edit & 161.50 & 143.26 & 60.39 & 17.25 \\
     Post-edit & 81.32 & 22.748 & 60.21 & 17.15 \\
     \hline
    \end{tabular}
    \caption{Our results for foreground and background color prediction for selected examples before and after manual editing.}
    \label{tab:manual_edit}
\end{table}

\begin{table}[t]
    \centering
    \begin{tabular}{ lcc } 
    \hline
    Ours vs & Mean preference rate & Std\\
     \hline
     Context-Aware \cite{DBLP:conf/iccv/Hou019} & 48.87\% & 0.15\\
     Samplenet \cite{DBLP:conf/cvpr/TangAOGA19} & 64.84\% & 0.19\\
     \hline
    \end{tabular}
    \caption{Results of the user study on the real world dataset \cite{DBLP:conf/cvpr/XuPCH17}}
    \label{tab:study}
\end{table}

\section{Conclusions}
In this work, we propose a novel method to estimate foreground and background colors given an initial alpha prediction. Our method is lightweight and can easily be used on top of any other alpha prediction method. We show that even initial alpha predictions that do not satisfy high-quality standards generate color predictions that are quantitatively better than the colors directly taken from the input image. We show through quantitative and qualitative evaluation that our method substantially outperforms the state-of-the-art in foreground color estimation. Further, the recurrent nature of our method allows users to manually edit parts of the candidate solutions with ease, which can propagate further and lead to better final predictions. We show that very rough edits to the background candidate solution can lead to a significantly better final foreground solution through minimal effort.

\section{Acknowledgements}
This publication has emanated from research conducted with the financial support of Science Foundation Ireland (SFI) under the Grant Number 15/RP/2776.

{\small
\bibliographystyle{ieee_fullname}
\bibliography{egbib}

\begin{thebibliography}{10}\itemsep=-1pt

\bibitem{DBLP:conf/cvpr/AksoyAP17}
Yagiz Aksoy, Tun{\c{c}}~Ozan Aydin, and Marc Pollefeys.
\newblock Designing effective inter-pixel information flow for natural image
  matting.
\newblock In {\em 2017 {IEEE} Conference on Computer Vision and Pattern
  Recognition, {CVPR} 2017, Honolulu, HI, USA, July 21-26, 2017}, pages
  228--236. {IEEE} Computer Society, 2017.

\bibitem{DBLP:journals/corr/ArjovskyCB17}
Mart{\'{\i}}n Arjovsky, Soumith Chintala, and L{\'{e}}on Bottou.
\newblock Wasserstein {GAN}.
\newblock {\em CoRR}, abs/1701.07875, 2017.

\bibitem{DBLP:journals/corr/abs-1909-04686}
Shaofan Cai, Xiaoshuai Zhang, Haoqiang Fan, Haibin Huang, Jiangyu Liu, Jiaming
  Liu, Jiaying Liu, Jue Wang, and Jian Sun.
\newblock Disentangled image matting.
\newblock In {\em Proceedings of the IEEE International Conference on Computer
  Vision}, pages 8819--8828, 2019.

\bibitem{DBLP:conf/cvpr/ChenLT12}
Qifeng Chen, Dingzeyu Li, and Chi{-}Keung Tang.
\newblock {KNN} matting.
\newblock In {\em 2012 {IEEE} Conference on Computer Vision and Pattern
  Recognition, Providence, RI, USA, June 16-21, 2012}, pages 869--876. {IEEE}
  Computer Society, 2012.

\bibitem{DBLP:conf/eccv/ChoTK16}
Donghyeon Cho, Yu{-}Wing Tai, and In{-}So Kweon.
\newblock Natural image matting using deep convolutional neural networks.
\newblock In Bastian Leibe, Jiri Matas, Nicu Sebe, and Max Welling, editors,
  {\em Computer Vision - {ECCV} 2016 - 14th European Conference, Amsterdam, The
  Netherlands, October 11-14, 2016, Proceedings, Part {II}}, volume 9906 of
  {\em Lecture Notes in Computer Science}, pages 626--643. Springer, 2016.

\bibitem{DBLP:conf/cvpr/ChuangCSS01}
Yung{-}Yu Chuang, Brian Curless, David Salesin, and Richard Szeliski.
\newblock A bayesian approach to digital matting.
\newblock In {\em 2001 {IEEE} Computer Society Conference on Computer Vision
  and Pattern Recognition {(CVPR} 2001), with CD-ROM, 8-14 December 2001,
  Kauai, HI, {USA}}, pages 264--271. {IEEE} Computer Society, 2001.

\bibitem{DBLP:conf/nips/GulrajaniAADC17}
Ishaan Gulrajani, Faruk Ahmed, Mart{\'{\i}}n Arjovsky, Vincent Dumoulin, and
  Aaron~C. Courville.
\newblock Improved training of wasserstein gans.
\newblock In Isabelle Guyon, Ulrike von Luxburg, Samy Bengio, Hanna~M. Wallach,
  Rob Fergus, S.~V.~N. Vishwanathan, and Roman Garnett, editors, {\em Advances
  in Neural Information Processing Systems 30: Annual Conference on Neural
  Information Processing Systems 2017, 4-9 December 2017, Long Beach, CA,
  {USA}}, pages 5767--5777, 2017.

\bibitem{DBLP:conf/iccv/Hou019}
Qiqi Hou and Feng Liu.
\newblock Context-aware image matting for simultaneous foreground and alpha
  estimation.
\newblock In {\em 2019 {IEEE/CVF} International Conference on Computer Vision,
  {ICCV} 2019, Seoul, Korea (South), October 27 - November 2, 2019}, pages
  4129--4138. {IEEE}, 2019.

\bibitem{DBLP:journals/corr/KingmaB14}
Diederik~P. Kingma and Jimmy Ba.
\newblock Adam: {A} method for stochastic optimization.
\newblock In Yoshua Bengio and Yann LeCun, editors, {\em 3rd International
  Conference on Learning Representations, {ICLR} 2015, San Diego, CA, USA, May
  7-9, 2015, Conference Track Proceedings}, 2015.

\bibitem{DBLP:conf/cvpr/LevinLW06}
Anat Levin, Dani Lischinski, and Yair Weiss.
\newblock A closed form solution to natural image matting.
\newblock In {\em 2006 {IEEE} Computer Society Conference on Computer Vision
  and Pattern Recognition {(CVPR} 2006), 17-22 June 2006, New York, NY, {USA}},
  pages 61--68. {IEEE} Computer Society, 2006.

\bibitem{DBLP:journals/corr/abs-2001-04069}
Yaoyi Li and Hongtao Lu.
\newblock Natural image matting via guided contextual attention.
\newblock {\em CoRR}, abs/2001.04069, 2020.

\bibitem{DBLP:conf/eccv/LinMBHPRDZ14}
Tsung{-}Yi Lin, Michael Maire, Serge~J. Belongie, James Hays, Pietro Perona,
  Deva Ramanan, Piotr Doll{\'{a}}r, and C.~Lawrence Zitnick.
\newblock Microsoft {COCO:} common objects in context.
\newblock In David~J. Fleet, Tom{\'{a}}s Pajdla, Bernt Schiele, and Tinne
  Tuytelaars, editors, {\em Computer Vision - {ECCV} 2014 - 13th European
  Conference, Zurich, Switzerland, September 6-12, 2014, Proceedings, Part
  {V}}, volume 8693 of {\em Lecture Notes in Computer Science}, pages 740--755.
  Springer, 2014.

\bibitem{DBLP:conf/bmvc/LutzAS18}
Sebastian Lutz, Konstantinos Amplianitis, and Aljosa Smolic.
\newblock Alphagan: Generative adversarial networks for natural image matting.
\newblock In {\em British Machine Vision Conference 2018, {BMVC} 2018,
  Northumbria University, Newcastle, UK, September 3-6, 2018}, page 259. {BMVA}
  Press, 2018.

\bibitem{DBLP:conf/iclr/MiyatoKKY18}
Takeru Miyato, Toshiki Kataoka, Masanori Koyama, and Yuichi Yoshida.
\newblock Spectral normalization for generative adversarial networks.
\newblock In {\em 6th International Conference on Learning Representations,
  {ICLR} 2018, Vancouver, BC, Canada, April 30 - May 3, 2018, Conference Track
  Proceedings}. OpenReview.net, 2018.

\bibitem{DBLP:conf/cvpr/PriceMC10}
Brian~L. Price, Bryan~S. Morse, and Scott Cohen.
\newblock Simultaneous foreground, background, and alpha estimation for image
  matting.
\newblock In {\em The Twenty-Third {IEEE} Conference on Computer Vision and
  Pattern Recognition, {CVPR} 2010, San Francisco, CA, USA, 13-18 June 2010},
  pages 2157--2164. {IEEE} Computer Society, 2010.

\bibitem{DBLP:journals/corr/PutzkyW17}
Patrick Putzky and Max Welling.
\newblock Recurrent inference machines for solving inverse problems.
\newblock {\em CoRR}, abs/1706.04008, 2017.

\bibitem{DBLP:conf/cvpr/RhemannRWGKR09}
Christoph Rhemann, Carsten Rother, Jue Wang, Margrit Gelautz, Pushmeet Kohli,
  and Pamela Rott.
\newblock A perceptually motivated online benchmark for image matting.
\newblock In {\em 2009 {IEEE} Computer Society Conference on Computer Vision
  and Pattern Recognition {(CVPR} 2009), 20-25 June 2009, Miami, Florida,
  {USA}}, pages 1826--1833. {IEEE} Computer Society, 2009.

\bibitem{DBLP:journals/corr/SimonyanZ14a}
Karen Simonyan and Andrew Zisserman.
\newblock Very deep convolutional networks for large-scale image recognition.
\newblock In Yoshua Bengio and Yann LeCun, editors, {\em 3rd International
  Conference on Learning Representations, {ICLR} 2015, San Diego, CA, USA, May
  7-9, 2015, Conference Track Proceedings}, 2015.

\bibitem{DBLP:conf/cvpr/TangAOGA19}
Jingwei Tang, Yagiz Aksoy, Cengiz {\"{O}}ztireli, Markus~H. Gross, and
  Tun{\c{c}}~Ozan Aydin.
\newblock Learning-based sampling for natural image matting.
\newblock In {\em {IEEE} Conference on Computer Vision and Pattern Recognition,
  {CVPR} 2019, Long Beach, CA, USA, June 16-20, 2019}, pages 3055--3063.
  Computer Vision Foundation / {IEEE}, 2019.

\bibitem{DBLP:conf/cvpr/XuPCH17}
Ning Xu, Brian~L. Price, Scott Cohen, and Thomas~S. Huang.
\newblock Deep image matting.
\newblock In {\em 2017 {IEEE} Conference on Computer Vision and Pattern
  Recognition, {CVPR} 2017, Honolulu, HI, USA, July 21-26, 2017}, pages
  311--320. {IEEE} Computer Society, 2017.

\bibitem{DBLP:journals/corr/abs-1806-03589}
Jiahui Yu, Zhe Lin, Jimei Yang, Xiaohui Shen, Xin Lu, and Thomas~S. Huang.
\newblock Free-form image inpainting with gated convolution.
\newblock {\em CoRR}, abs/1806.03589, 2018.

\bibitem{DBLP:conf/cvpr/Yu0YSLH18}
Jiahui Yu, Zhe Lin, Jimei Yang, Xiaohui Shen, Xin Lu, and Thomas~S. Huang.
\newblock Generative image inpainting with contextual attention.
\newblock In {\em 2018 {IEEE} Conference on Computer Vision and Pattern
  Recognition, {CVPR} 2018, Salt Lake City, UT, USA, June 18-22, 2018}, pages
  5505--5514. {IEEE} Computer Society, 2018.

\bibitem{DBLP:conf/cvpr/ZhangGFRHBX19}
Yunke Zhang, Lixue Gong, Lubin Fan, Peiran Ren, Qixing Huang, Hujun Bao, and
  Weiwei Xu.
\newblock A late fusion {CNN} for digital matting.
\newblock In {\em {IEEE} Conference on Computer Vision and Pattern Recognition,
  {CVPR} 2019, Long Beach, CA, USA, June 16-20, 2019}, pages 7469--7478.
  Computer Vision Foundation / {IEEE}, 2019.

\end{thebibliography}
}

\end{document}